\pdfoutput=1
\documentclass[11pt]{article}

\usepackage[preprint]{acl}

\usepackage{times}
\usepackage{latexsym}
\usepackage{dashrule}

\usepackage[T1]{fontenc}
\usepackage[utf8]{inputenc}

\usepackage{microtype}

\usepackage{inconsolata}

\usepackage{graphicx}

\usepackage{hyperref}
\usepackage{url}            % simple URL typesetting
\usepackage{booktabs}       % professional-quality tables
\usepackage{amsfonts}       % blackboard math symbols
\usepackage{nicefrac}       % compact symbols for 1/2, etc.
\usepackage{xcolor}         % colors
\usepackage{float}
\usepackage{multirow}
\usepackage{amsthm}
\usepackage{xcolor}
\usepackage{comment}
\usepackage{mathtools}
\usepackage[most]{tcolorbox}
\usepackage[shortlabels]{enumitem}

\usepackage{todonotes}
\usepackage{colortbl}
\usepackage{enumitem}
\usepackage{bm} 
\usepackage{hwemoji}
\usepackage{pifont}

\usepackage{url}
\usepackage{hyperref}
\usepackage{array}
\usepackage{pifont}
\usepackage{tabularx}
\usepackage{adjustbox}
\usepackage{multirow}
\usepackage{enumitem}
\usepackage{xspace}
\usepackage{tcolorbox}
\usepackage{subcaption}
\usepackage{booktabs,amsfonts,dcolumn}
\usepackage{amsmath,amsthm,amsfonts,amssymb,bm,stmaryrd,bbm}
\usepackage{makecell}
\usepackage{graphicx}
\usepackage{colortbl}
\usepackage{color,soul}
\usepackage{wrapfig}
\usepackage{float}
\usepackage{arydshln}

\newcolumntype{P}[1]{>{\centering\arraybackslash}p{#1}}
\newcommand{\mydots}{\hbox to 1em{.\hss.\hss.}}

\newcommand{\pz}{\phantom{0}}

\renewcommand{\paragraph}[1]{\vspace{0.2cm}\noindent\textbf{#1}}

\definecolor{ccon}{HTML}{fee9d4}
\definecolor{cood}{HTML}{d8f0d3}
\definecolor{cid}{HTML}{dae8f5}

\definecolor{gred}{HTML}{cc0200}
\definecolor{ggreen}{HTML}{38761c}

\definecolor{c1}{cmyk}{0,0.6175,0.8848,0.1490}
\definecolor{c2}{cmyk}{0.1127,0.6690,0,0.4431}
\definecolor{c3}{cmyk}{0.3081,0,0.7209,0.3255}
\definecolor{c4}{cmyk}{0.6765,0.2017,0,0.0667}
\definecolor{c5}{cmyk}{0,0.8765,0.7099,0.3647}

\newtcbox{\hlprimarytab}{on line, rounded corners, box align=base, colback=green!60,colframe=white,size=fbox,arc=3pt, before upper=\strut, top=-2pt, bottom=-4pt, left=-2pt, right=-2pt, boxrule=0pt}
\newtcbox{\hlsecondarytab}{on line, box align=base, colback=red!45,colframe=white,size=fbox,arc=3pt, before upper=\strut, top=-2pt, bottom=-4pt, left=-2pt, right=-2pt, boxrule=0pt}

\newcommand{\da}[1]{(\textcolor{magenta}{$\downarrow$ #1})}
\newcommand{\ua}[1]{(\textcolor{teal}{$\uparrow$ #1})}

\makeatletter
\def\thickhline{%
  \noalign{\ifnum0=`}\fi\hrule \@height \thickarrayrulewidth \futurelet
   \reserved@a\@xthickhline}
\def\@xthickhline{\ifx\reserved@a\thickhline
               \vskip\doublerulesep
               \vskip-\thickarrayrulewidth
             \fi
      \ifnum0=`{\fi}}
\makeatother
\newlength{\thickarrayrulewidth}
\title{Do Robot Snakes Dream like Electric Sheep? \\ Investigating the Effects of Architectural Inductive Biases on Hallucination}

\author{%
    Jerry Huang \\
    Chandar Research Lab \\
    Mila \& Universit\'{e} de Montr\'{e}al
    \And
    Prasanna Parthasarathi\thanks{Corresponding author: \href{mailto:pp1403@gmail.com}{\texttt{pp1403@gmail.com}}} \\
    Noah's Ark Lab
    \And
    Mehdi Rezagholizadeh\thanks{Work done while at Noah's Ark Lab.} \\
    Advanced Micro Devices
    \AND
    Boxing Chen \\
    Noah's Ark Lab
    \And
    Sarath Chandar \\
    Chandar Research Lab \\
    Mila \& Polytechnique Montr\'{e}al \\ 
    CIFAR AI Chair
}

\begin{document}

\maketitle

\begin{abstract}
% The use of large language models have brought about various improvements in recent years, through the use of new data, hardware advances, as well as new architectures meant to overcome limitations of previous models.

% While the growth in prominence of large language models in everyday life has validated their skills as a contemporary renaissance man, their increasing presence have led to simultaneous concerns about their tendency to \textit{hallucinate} false or misleading information.
% Not connected here, will think of a way to do that later.
% Concurrently, an increasing focus on the computational limitations associated the traditional Transformer-based structure LLMs have brought about new alternatives, in particular sequence models, meant to overcome them. 

% Prominence of LLMs in everyday life owing to their generative abilities also entails some risk and costly mitigation.

The growth in prominence of large language models (LLMs) in everyday life can be largely attributed to their generative abilities, yet some of this is also owed to the risks and costs associated with their use. On one front is their tendency to \textit{hallucinate} false or misleading information, limiting their reliability. On another is the increasing focus on the computational limitations associated with traditional self-attention based LLMs, which has brought about new alternatives, in particular recurrent models, meant to overcome them. Yet it remains uncommon to consider these two concerns simultaneously. Do changes in architecture exacerbate/alleviate existing concerns about hallucinations? Do they affect how and where they occur? Through an extensive evaluation, we study how these architecture-based inductive biases affect the propensity to hallucinate. While hallucination remains a general phenomenon not limited to specific architectures, the situations in which they occur and the ease with which specific types of hallucinations can be induced can significantly differ based on the model architecture. These findings highlight the need for better understanding both these problems in conjunction with each other, as well as consider how to design more universal techniques for handling hallucinations. %\todo{Need to update the parts about what we find/observe based on the way we end the intro.}
\end{abstract}

\section{Introduction}

Large language models (LLMs) have rapidly emerged as a every-day tool in modern life~\citep{openai2024gpt4}, with many relying on their abilities to accomplish a variety of specific tasks. However, this opened up concerns relating to their propensity to \textit{hallucinate}~\citep{huang2023survey}, with no concrete reasons for this behaviour~\citep{dziri2022origin, rawte2023troubling, chen2024bigger}, hindering the ability to directly train LLMs that are consistently factual or able to explain themselves through their factual knowledge~\citep{madsen2024selfexplanations, prato-etal-2023-epik, prato-etal-2024-large, toolscl}. 

In parallel, as LLMs evolve and existing limitations are discovered, alternative architectures have become increasingly common and grow in popularity. In particular, Transformer LLMs~\citep{vaswani2017attention} and linear sequence models~\citep{gu2022efficiently, gu2024mamba} present contrasting methods of encoding sequences, with the Transformer using attention~\citep{bahdanau2016neural} to form lossless representations of the context, while linear sequence models follow the recurrent neural network~\citep{rumelhart1987parallel, jordan1986serial} in their use of a compressed state representation.

%Yet while methods aimed towards forestalling such events occurs post-training, such as through fine-tuning with human feedback~\citep{lightman2023lets} or during inference through methods such as proving external knowledge bases and self-reflection~\citep{lee2022factuality, gao2023rarr, manakul2023selfcheckgpt} have proven useful, it remains under-explored how hallucinations are the result of inherent architectural biases. 
%While prior work~\citep{li2022pre} suggested that LLMs simply encode structure through the next-token-prediction objective and thus fail to encode truly meaningful understanding of the encoded knowledge~\citep{kadavath2022language}, hinting at the potential connection between training objectives and hallucination, further understanding of this requires uncovering the links between the phenomena at hand (hallucination) and the inductive biases presented by the models we use. 

With the intensifying focus in both directions, a notable void exists in verifying how each can affect the others in conjunction. For example, existing works in hallucination detection and mitigation focus almost exclusively on Transformer-based models~\citep{maynez2020faithfulness, longpre2021entity, guerreiro2023looking, shi2023large, ji2023rho, farquhar2024hallucinations, wei2024measuring}, without extension to recurrent-style models, despite the use of a unified hidden representation potentially acting as an information bottleneck that can induce more common hallucinations. This lack of unified understanding on both topics prompts the need for a more explicit investigation.
% Based on the recurrent neural nework~\citep{rumelhart1987parallel, jordan1986serial}, these models use a compressed hidden state to encode information, with specific parameterizations~\citep{gu2020hippo} that enable the memorization of complete sequences and the elimination of gradient scaling~\citep{bengio1994learning, hochreiter2001gradient, pascanu2013difficulty}. 
% While the use of a lossy encoding~\citep{vardasbi2023state} would suggest greater difficulty extracting context-specific factual information and subsequently more common hallucinations, the inductive bias of such models inherently enable these models to properly model relationships between different parts of the sequence. A more explicit investigation is thus necessary.

In this work, we comprehensively explore the differences between pure attention LLMs and recurrent LLMs, specifically with respect to the propensity to hallucinate. 
Using a set of 20 different hallucination tasks, categorized into 6 groups that evaluate both faithfulness and factuality hallucinations, we evaluate across numerous open-source LLMs that range in scale from under 1B parameters to 70B parameters all the while covering a variety of different architecture choices such as self-attention, recurrent and hybrid models. We further evaluate across factors such as instruction-tuning, all to build a more comprehensive picture of architecture-specific phenomena with respect to hallucination. From this, we observe the following:%\todo{This needs some more specifics}
\begin{enumerate}[itemsep=0ex, leftmargin=0.5cm, label=(\arabic*)]
    \item Viewed very broadly over various different tasks and settings, neither Transformer-based nor recurrent/hybrid LLMs appear to induce hallucinations more often than others.
    \item However, shifting to individual tasks, it becomes evident that they result in disparate tendencies on specific tasks evaluating for unique criteria such as recalling long-tailed factual knowledge and falling into memorization traps, highlighting that model architecture may promote specific behavior that renders some types of hallucinations more common.
    \item Evaluating the effects of scale and instruction-tuning, we observe that though factuality remains dependent on model size, recurrent/hybrid architectures are often more faithful at smaller sizes and observe significantly fewer faithfulness benefits from instruction-tuning and scaling compared to self-attention models.
\end{enumerate}
These results highlight that while some hallucinations may be quite consistent across architectures, others are the direct results of specific model design choices that go into the LLM construction. This hints towards the need for more careful consideration on this front, as different techniques for addressing this problem can potentially be highly catered towards specific models, bringing to the forefront the need for better consideration of both these problems in the face of each other.

\section{Related Work}

\paragraph{Hallucinations in LLMs.} \textit{Hallucination} broadly refers to when LLMs generate information that does not directly follow from the context, such as nonsensical or irrelevant answers to questions~\citep{ji2023survey}. While it has grown in importance due to its direct relationship with ensuring the safe and responsible use of LLMs, both classifying and quantifying the hallucinations is challenging. In particular, it is difficult to ascertain if the divergence occurs because of specific data heuristics~\citep{lebret2016-neural, wiseman2017-challenges} or because of the innate lack of similarity between pre-training and downstream tasks~\citep{rashkin2021-increasing}, while measuring for hallucinations automatically introduces various biases~\citep{reiter-2018-structured, tian2020sticking, ganguli2022red} that may not fully capture the scope of the errors. However, a variety of task-specific benchmarks~\citep{li2020-slot, pagnoni2021-understanding, zhou2021-detecting, santhanam2022rome} have shown various LLMs to struggle with factual inconsistencies, highlighting a need to render them safer for every-day use.

\paragraph{LLM Architectures.} General LLM architecture can be broadly thought to be composed of two components: token mixers which serves to model transformations between time steps (such as attention and recurrent layers) and channel mixers, such as multi-layer perceptrons and mixtures-of-experts (MoE)~\citep{jacobs1991adaptive, shazeer2017outrageouslylargeneuralnetworks}, which allow communication between different channels within a single time step. Accordingly, token mixing often forms a computational bottleneck in terms of time complexity while channel mixers consist of a memory bottleneck. Though the contemporary standard for token mixing remains self-attention, alternatives are becoming increasingly common as they begin to display more promise. These include the use of linear attention~\citep{katharopoulos2020transformers} to compensate for the quadratic memory complexity of vanilla self-attention, to new recurrent models~\citep{gu2024mamba, botev2024recurrentgemma} that function as a linear recurrent neural network but can process all elements of a sequence in parallel, and well as hybrid mixtures of recurrent mechanisms and attention~\citep{jamba, dao2024transformers} that have emerged as a meaningful competitor.

\paragraph{Architecture and Hallucination.} Despite growing research in new architectural components, their effects on hallucination have yet to be studied. While some works~\citep{madsen2024faithfulness, hu2024mitigatinglargelanguagemodel, schimanski2024-towards} have proposed modifications to either the learning/generation pipeline as a way of reducing hallucinations, proposals for hallucination reduction through structural modification have yet to be suggested. Additionally, though some works~\citep{elhage2021mathematical, fu2023monarch, lutati2023focus, poli2024mechanisticdesignscalinghybrid} have demonstrated self-attention to aptly solve synthetic tasks that form an essential component of language modeling, these fail to remain faithful on more realistic datasets, punctuating a major limitation of existing LLMs. Finally, though tangential work demonstrates that recurrent models may suffer from issues with information retention~\citep{vardasbi2023state}, formally defining a link with hallucination remains necessary. These issues lead to a lack of clarity on this front, accentuating the need for a formal investigation comparing hallucination alongside architectural paradigms. In particular, the choice of token mixer is important, as this is the component that governs how information is shared between different elements of the sequence to form a holistic representation. As such, an effective token mixer will adequately enable enough information from the context to propagate forward during the generation phase~\citep{olsson2022incontext, arora2024zoology}, potentially enabling models to hallucinate less.

\section{Background}\label{sec:background}

\paragraph{Attention for Sequences.} Vanilla self-attention as used in Transformers is powerful but costly. When provided an embedded text representation as a sequence of tokens $\bm{x}\in\mathbb{R}^{L \times d}$, each Transformer layer in the network applies a function
\begin{equation}
    T_\ell(\bm{x}) = \text{FF}_\ell(A_\ell(\bm{x}) + \bm{x}) + A_\ell(\bm{x})
    \label{eq:transformer_layer}
\end{equation}
where $A_\ell$ is the self-attention of the $\ell$-th layer and $\text{FF}_\ell$ is the following feed-forward network. Self-attention computes, for a token at position $i$ in a sequence, a weighted average of the feature representations of all tokens (the values $\bm{V}_\ell$) in the sequence with a weight proportional to a similarity score between $i$ (the query $\bm{Q}_\ell$ at position $i$) and the rest of the sequence (the keys $\bm{K}_\ell$). In particular, these can be computed for all positions in parallel
\begin{equation}\label{eq:attention}
    \begin{split}
    &\bm{Q}_\ell = \bm{x}\bm{W}_\ell^{\bm{Q}} \quad \bm{K}_\ell = \bm{x}\bm{W}_\ell^{\bm{K}} \quad \bm{V}_\ell = \bm{x}\bm{W}_\ell^{\bm{V}} \\
    &A_\ell(\bm{x}) = \bm{V}'_\ell = \text{softmax}\big({\bm{Q}_\ell\bm{K}_\ell^T}/{\sqrt{d}}\big)\bm{V}_\ell
    \end{split}
\end{equation}
providing the model a lossless representation of the complete past context. This can be seen as equivalent to search and retrieval within a database, where search is defined using query-key parameterizations and retrieval with value parameterization. In this setting, the database from which the information is being retrieved is equivalent to the model parameters, which store information from a training corpus that the weights $\bm{W}^{\bm{K}}, \bm{W}^{\bm{Q}}, \bm{W}^{\bm{V}}$ parameterizing the model attempt to mimic.

Multi-head attention, a variant of self-attention popularized by the Transformer~\citep{vaswani2017attention}, has become the dominant variant used in LLMs such as LLaMA~\citep{touvron2023llama} and Gemma~\citep{gemma_2024}. Additional variants, such as multi-query attention from Falcon~\citep{almazrouei2023falconseriesopenlanguage} and grouped-query attention from Mistral~\citep{jiang2023mistral}, adapt multi-head attention but remain build upon the same underlying principle of self-attention. Consequently, such methods are considered to fall under the family of self-attention token mixers.
% The primary bottleneck concerns the softmax operation, which operates in $O(L^2)$ time when applied naively. As such, modifications have been made to render this more efficient~\citep{linearattention, peng2021random, schlag2021linear, nystromformer}, yet the problem remains unresolved.

\paragraph{Recurrent LLMs.} A concern for Transformers is the quadratic complexity of attention with sequence length, leading a focused on improving this bound as it directly affects the ability to learn from long sequences. Instead of directly modifying attention~\citep{katharopoulos2020transformers, Kitaev2020Reformer, choromanski2021rethinking, zeng2025zeta}, \citet{gu2020hippo} motivated a novel paradigm using state-space models (SSMs) from control theory. SSMs map an input $x(t)\in\mathbb{R}^d$ to an intermediate state $h(t)\in\mathbb{R}^n$ that is then projected to an output $y(t)\in\mathbb{R}^d$:
\begin{equation*}\label{eq:ssm_cont}
    h'(t) = {\bm{A}}h(t) + {\bm{B}}x(t), \quad y(t) = {\bm{C}}h(t) + {\bm{D}}x(t)
\end{equation*}
where $\bm{A}$, $\bm{B}$, $\bm{C}$ and $\bm{D}$ are trainable parameters and $h'(t)$ represents the rate at which $h(t)$ changes. \citet{gu2021combining} use this paradigm to define a recurrent model to work on discrete signals, in which case the input can be regarded as discretized data sampled from a continuous signal with a step size $\Delta$, for which the corresponding SSM is defined by:
\begin{equation*}\label{eq:ssm}
    \begin{split}
        h_{t} &= \overline{\bm{A}}h_{t-1} + \overline{\bm{B}}x_{t} \quad y_{t} = \overline{\bm{C}}h_{t} + \overline{\bm{D}}x_{t} \\
        \overline{\bm{A}} &= \frac{\big(I+{\Delta}\bm{A}/{2})}{\big(I-{\Delta}\bm{A}/{2}\big)} \quad \overline{\bm{B}} = \frac{\Delta\bm{B}}{\big(I-{\Delta}\bm{A}/{2}\big)}
    \end{split}
\end{equation*}
and $\overline{\bm{C}} = \bm{C}$ ($\overline{\bm{D}}$ is equivalent to a residual connection and set to $\bm{0}$.) Thus
\begin{equation*}\label{eq:ssm_kernel}
    % \begin{split}
    %     \overline{\bm{K}} &= (\overline{\bm{CB}}, \overline{\bm{CAB}}, \dots, \overline{\bm{CA}}^{L-1}\overline{\bm{B}}) \\
    %     \bm{y} &= \sum_{j=0}^{L-1} \overline{\bm{CA}}^{j}\overline{\bm{B}} x_{L-j}= \overline{\bm{K}} * \bm{x}
    % \end{split}
        \begin{split}
        \overline{\bm{K}} &= (\overline{\bm{CB}}, \overline{\bm{CAB}}, \dots, \overline{\bm{CA}}^{L-1}\overline{\bm{B}}) \quad \bm{y} = \overline{\bm{K}} * \bm{x}
    \end{split}
\end{equation*}
where $\overline{\bm{K}}$ is the SSM kernel. As $\bm{y}$ can be computed in $O(L\log L)$ with a Fast Fourier Transform~\citep{algorithms}, the entire output can be computed in tandem, given the matrices that parameterize the system. Furthermore, setting $\bm{A}$ as a Hurwitz matrix, SSMs can preserve long-term information, overcoming a long-standing issue~\citep{bengio1994learning} with prior recurrent models~\citep{lstm, cho2014-learning}. 

Further works have modified this structure; \citet{gu2024mamba} use input-dependent $\bm{B}$ and $\bm{C}$ information filtering. LRU/Hawk~\citep{lru, de2024griffin} remove the discretization step from a continuous signal and instead learn discrete matrices for $\bm{A}$, $\bm{B}$, $\bm{C}$ and $\bm{D}$ directly. RWKV~\citep{peng2023rwkv, peng2024eaglefinchrwkvmatrixvalued} uses a novel WKV sequence mixing operator that acts like an RNN. However, these models all share the use of hidden states from which token-level information must pass through in order to interact with future tokens. This has also led to the creation of hybrid mixtures of recurrence and attention~\citep{dao2024transformers, de2024griffin, jamba} to form new classes of models meant to capture the benefits of both.

Different sequence (or time) mixing modules attempt to encode the history of tokens in a fixed representation. Recurrent architectures maintain this representation and learn to update it during pre-training, but this can fall short of attention~\citep{vardasbi2023state, jelassi2024repeat, huang-2025-well, resona} as they are constrained by its size~\citep{arora2024zoology}. Meanwhile, attention has a theoretically infinite context window (albeit being costly). As recurrent structures are constrained to make the predictions based on the recent context (which in practice can be long, but remains shorter than what self-attention can attend to), the expressivity of the hidden representation directly affects task performance~\citep{sun2024learninglearntesttime} and parametric knowledge can be predisposed to favor the prediction of certain tokens. We model this effect on tasks under the lens of hallucination: faithfulness and factuality. Faithfulness expects the model to follow an instruction (that is largely recent), while factuality assumes the use of token-to-token relations learned during pre-training for a broader contextual representation. Following this, we ask two questions: (1) Do self-attention and recurrent architectures hallucinate differently? (2) Does instruction-tuning and/or size benefit the different classes of LLMs equally?

\begin{figure*}
    \centering
    \includegraphics[width=\linewidth]{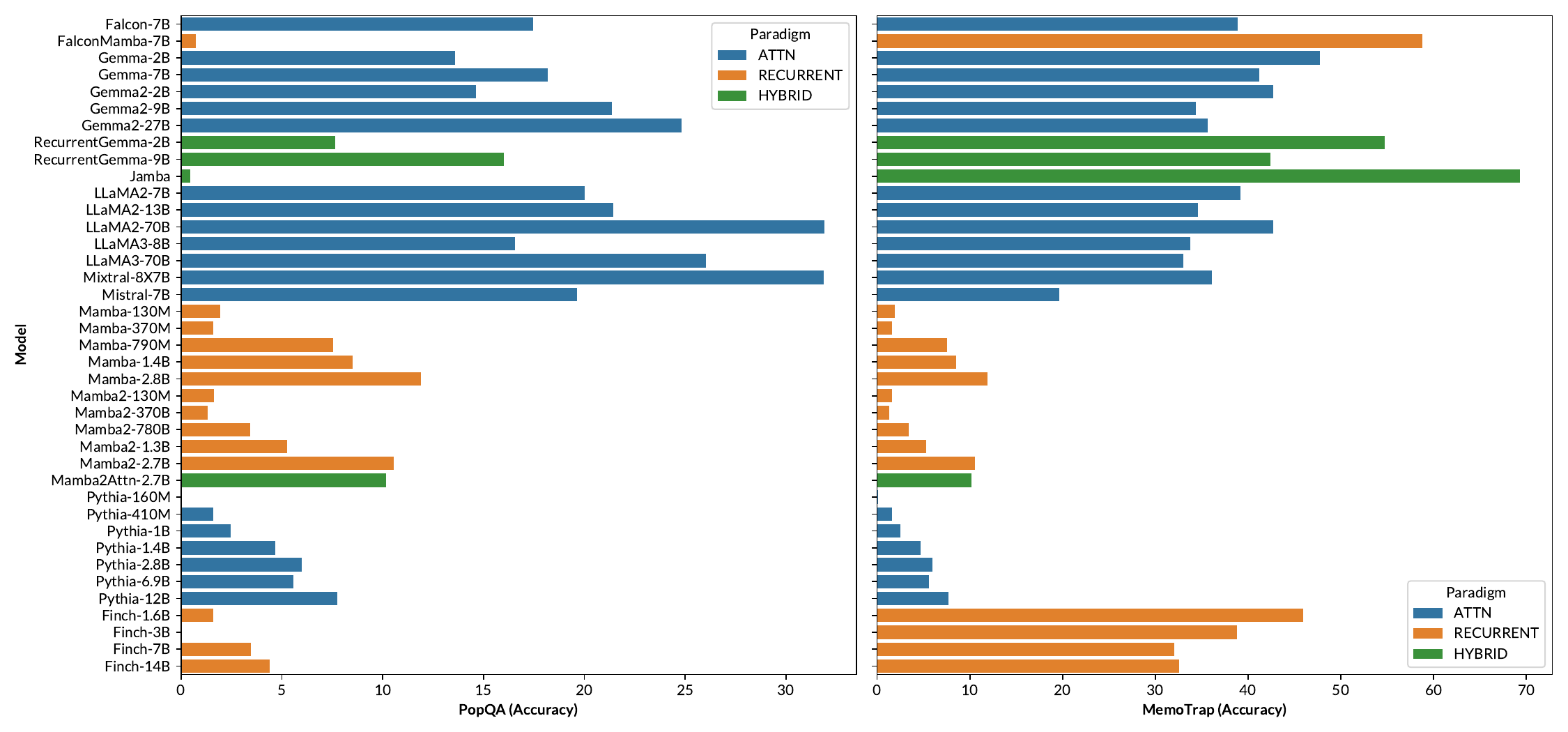}
    \caption{Performance on \textsc{PopQA} (left) and \textsc{MemoTrap} (right). Recurrent and hybrid models significantly outperform similar pure attention-based alternatives on \textsc{MemoTrap}, but the opposite is true on \textsc{PopQA}.}
    \label{fig:o12}
\end{figure*}

\section{Experiments and Results}

\paragraph{Models.} To investigate, we use models that vary in size and architecture, with a particular focus on different sequence mixing methods.
These include self-attention models (\texttt{Pythia}, \texttt{LLaMA2/3}~\citep{touvron2023llama}, \texttt{Falcon}~\citep{almazrouei2023falconseriesopenlanguage}, \texttt{Mistral}~\citep{jiang2023mistral}, \texttt{Gemma}~\citep{gemma_2024} and \texttt{Mixtral}~\citep{jiang2024mixtralexperts}), recurrent models  (\texttt{Mamba}~\citep{gu2024mamba, dao2024transformers}, \texttt{FalconMamba} and \texttt{RWKV}/\texttt{Finch}~\citep{peng2023rwkv}), as well as hybrid models (\texttt{RecurrentGemma}~\citep{botev2024recurrentgemma}, \texttt{Jamba}~\citep{jamba}). We use base and instruction-tuned variants when available. Additional details are available in \autoref{app:technical_details}.

\paragraph{Datasets.} We evaluate on the Hallucination Leaderboard~\citep{hong2024hallucinationsleaderboardopen}, consisting of tasks
\begin{enumerate}[itemsep=0ex, leftmargin=0.5cm, label=\arabic*)]
    \item \textbf{Closed-book Open-domain QA}$^{\color{magenta}\clubsuit}$: 
        \textsc{NQ-Open}~\citep{kwiatkowski2019-natural}, 
        \textsc{TriviaQA}~\citep{joshi2017-triviaqa}, 
        \textsc{TruthfulQA}~\citep{lin2022truthfulqa}, 
        \textsc{PopQA}~\citep{mallen2023trust}
    \item \textbf{Summarization}$^\spadesuit$: 
        \textsc{XSum}~\citep{Narayan2018DontGM},    
        \textsc{CNN/DM}~\citep{see2017-get}
    \item \textbf{Reading Comprehension}$^\spadesuit$: 
        \textsc{RACE}~\citep{lai2017-race}, 
        \textsc{SQuADv2}~\citep{rajpurkar2018-know}, 
        \textsc{NQ-Swap}~\citep{longpre2021-entity}
    \item \textbf{Instruction Following}$^\spadesuit$: 
        \textsc{MemoTrap}~\citep{liu2023memotrap}, 
        \textsc{IFEval}~\citep{zhou2023instructionfollowing}
    \item \textbf{Hallucination Detection}: 
        \textsc{FaithDial}$^\spadesuit$~\citep{dziri2022faithdial}, 
        \textsc{HaluEval}$^\spadesuit$~\citep{li2023halueval}, 
        \textsc{True-False}$^{\color{magenta}\clubsuit}$~\citep{azaria-mitchell-2023internal}
    \item \textbf{Fact Checking}$^{\color{magenta}\clubsuit}$: 
        \textsc{FEVER}~\citep{thorne2018-fever}
\end{enumerate}
% This enables us to evaluate effects along different axes, in particular within specific datasets, domains or even related concepts. Each evaluation uses different evaluation metrics listed in their respective results columns. 
For tasks, a higher score (ranging from 0 to 100) indicates better performance. Tasks are further divided into \textit{faithfulness} ($\spadesuit$), i.e. whether the generation adheres to the given context, and \textit{factuality} (${\color{magenta}\clubsuit}$), i.e. whether the generation is factually correct.

\subsection{Investigating Task Biases} Our first interest is to verify whether specific choices in architecture can lead to highly evident patterns in performance\footnote{Numbers provided in \autoref{tab:all}, \ref{tab:categories} and \ref{tab:group} in \autoref{app:complete_tables}.}.
%This is particularly interesting for the task of summarization, as previous work~\citep{vardasbi2023state} has suggested that recurrent models can struggle with an inability to compress information into a hidden state, something that is much more relevant in this specific scenario.
While direct differences can be difficult to quantify due to differences in how each model is trained, some pattern are consistent, described as follows.

% Looking more closely at task classes, however, this becomes less the case. In particular, recurrent models would appear to underperform attention-based models on summarization tasks. For example, while RecurrentGemma models appear to overall slightly achieve lower scores compared to 	exttt{Gemma} models (of both generations). Meanwhile, both Finch/RWKV and Mamba models underperform similarly sized 	exttt{Pythia} models despite . FalconMamba meanwhile slightly perform only slightly better than Falcon despite training on over 3 times the number of tokens. This relates to the findings of previous work~\citep{vardasbi2023state}, which have suggested that recurrent models can struggle with an inability to summarize a source sentence in a single hidden state, something that is much more relevant in tasks such as summarization and machine translation compared to the other explored tasks. In particular the longer sequences within the summarization tasks leads to denser hidden states, making it more difficult for the model to extract a holistic understanding of the context from it. As the other tasks categories generally generally use significantly shorter sequences or do not require explicitly needing to retain information from a context, this could mean that the hidden state has the potential to lead to increased hallucination for longer contexts.

% \begin{tcolorbox}[
%     colback=red!5!white,
%     colframe=red!75!black,
%     left*=2mm,
%     right*=2mm,
%     halign=center
% ]

\begin{center}
    % \fbox{
    \textbf{O1: Sequence models miss rare knowledge.}
    % }
\end{center}
% \end{tcolorbox}
Interestingly, recurrent and hybrid models all significantly underwhelm on \textsc{PopQA}, a task that tests for uncommon knowledge (left plot in \autoref{fig:o12}). For example, \texttt{FalconMamba} achieves an exact-match (EM) score of 0.7 compared to 17.5 by a similar \texttt{Falcon}. Similarly, 2B and 9B \texttt{RecurrentGemma} achieve 7.6 and 16.0 EM compared to 14.6/21.3 EM for 2B/7B \texttt{Gemma} and 13.6/18.2 by 2B/9B \texttt{Gemma2}, while \texttt{Jamba} (0.4 EM) performs worse than a similarly sized \texttt{Mixtral} (31.9 EM).

Previous work~\citep{vardasbi2023state} suggests that hidden states can become dense and difficult to extract information from. However, the brevity and simplicity of the prompt here begs the question whether the hidden representation has the opportunity to become significantly dense. Additional works~\citep{dao2024transformers} instead posit that using recurrent models with additional mechanisms enabling input filtering can enhance performance on domains such as language; however, the relevance of this claim here is again questionable. Yet this result is significant as it indicates recurrent models, despite the task's simplicity, struggle when compared to attention-based equivalents, suggesting a link with the sequence mixing. Instead, the reason may be that recurrent/hybrid models do not learn by explicitly memorizing and acting as a retrieval system (as discussed in \autoref{sec:background}), where the learning mechanism may not have led to said information being stored within parameters. %In this sense, despite potentially observing low-tail information, the learning mechanism may not have led to said information being stored within parameters, making recurrent/hybrid models more likely to produce incorrect information in such settings with higher degrees of uncertainty.

\begin{table*}[ht!]
    \centering
    \resizebox{\linewidth}{!}{
    % \begin{tabular}{lrrrrrr}
    \begin{tabular}{lcccccccc}
        \toprule
        \textbf{Model Name} & \textbf{Hallu. Detection} & \textbf{Instr. Following} & \textbf{Closed-Book QA} & \textbf{Reading Comp.} & \textbf{Sum.} & \textbf{Fact-Checking} & \textsc{Faithfulness} & \textsc{Factuality} \\
        \midrule
        \multicolumn{9}{c}{\textit{Attention-Only Models}} \\
        \midrule
        \textbf{Gemma-2B} & 49.65 \ua{\tt11.51} & 30.25 \ua{\tt13.84} & 38.20 \da{\tt-6.92} & 38.96 \da{\tt-13.96} & 24.93 \da{\tt-4.53} & 54.14  \ua{\tt\pz 1.60} & 35.85 \ua{\tt\pz 1.92} & 32.56 \da{\tt-4.56} \\
        \textbf{Gemma-7B} & 53.81 \da{\tt-5.67} & 31.94 \ua{\tt\pz 6.79} & 27.43 \da{\tt-4.13} & 31.35 \da{\tt-5.83} & 20.63 \ua{\tt\pz 1.04} & 43.28  \da{\tt-5.22} & 36.48 \ua{\tt\pz 2.71} & 44.50 \da{\tt-6.04} \\
        \midrule
        \textbf{Gemma2-2B} & 58.33 \ua{\tt\pz 0.64} & 26.73 \ua{\tt10.90} & 29.59 \ua{\tt\pz 4.83} & 32.95 \da{\tt-5.26} & 16.04 \da{\tt-0.47} & 39.57  \ua{\tt18.96} & 35.61 \ua{\tt\pz 0.65} & 34.96 \da{\tt-0.98} \\
        \textbf{Gemma2-9B} & 65.64 \ua{\tt\pz 6.31} & 23.76 \ua{\tt30.07} & 39.36 \ua{\tt\pz 3.13} & 39.74 \da{\tt-0.70} & 21.38 \da{\tt-3.35} & 62.06  \ua{\tt\pz 6.47} & 40.56 \ua{\tt\pz 7.55} & 46.85 \ua{\tt\pz 3.13} \\
        \textbf{Gemma2-27B} & 62.03 \ua{\tt12.43} & 26.62 \ua{\tt33.58} & 47.19 \ua{\tt\pz 2.86} & 43.04 \ua{\tt\pz 5.02} & 28.92 \da{\tt-0.74} & 68.28  \ua{\tt\pz 1.00} & 41.54 \ua{\tt13.02} & 53.92 \ua{\tt\pz 2.36} \\
        \midrule
        \textbf{LLaMA2-7B} & 53.63 \ua{\tt\pz 4.30} & 28.94 \ua{\tt\pz 8.99} & 37.64 \da{\tt-1.89} & 27.58 \ua{\tt\pz 3.74} & 25.19 \ua{\tt\pz 0.69} & 51.38  \ua{\tt\pz 5.88} & 35.14 \ua{\tt\pz 4.01} & 42.51 \da{\tt-0.12}\\
        \textbf{LLaMA2-13B} & 67.80 \da{\tt-3.68} & 26.57 \ua{\tt\pz 9.72} & 39.65 \da{\tt-0.65} & 32.17 \da{\tt-3.64} & 27.33 \da{\tt-0.23} & 62.35  \da{\tt-1.64} & 40.83 \ua{\tt\pz 0.88} & 46.27 \da{\tt-0.49}\\
        \textbf{LLaMA2-70B} & 62.34 \ua{\tt10.78} & 30.24 \ua{\tt17.11} & 47.85 \da{\tt-2.53} & 39.48 \da{\tt-8.21} & 28.00 \da{\tt-0.55} & 66.63  \da{\tt-1.45} & 41.74 \ua{\tt\pz 5.82} & 53.69 \da{\tt-2.34}\\
        \midrule
        \textbf{LLaMA3-8B} & 60.76 \ua{\tt10.12} & 22.06 \ua{\tt26.76} & 41.55 \ua{\tt\pz 1.30} & 33.52 \ua{\tt\pz 3.62} & 26.62 \da{\tt-1.60} & 60.84  \ua{\tt\pz 4.82} & 37.60 \ua{\tt10.09} & 48.14 \ua{\tt\pz 1.60}\\
        \textbf{LLaMA3-70B} & 71.78 \ua{\tt\pz 8.69} & 21.59 \ua{\tt25.60} & 48.07 \ua{\tt\pz 3.43} & 46.38 \da{\tt-7.73} & 28.91 \da{\tt-1.68} & 69.57  \ua{\tt\pz 0.99} & 45.92 \ua{\tt\pz 6.23} & 54.80 \ua{\tt\pz 2.81}\\
        \midrule
        \textbf{Mistral-7B} & 60.48 \ua{\tt\pz 3.36} & 28.39 \ua{\tt16.85} & 41.24 \ua{\tt\pz 5.21} & 32.03 \ua{\tt\pz 1.92} & 26.77 \da{\tt-0.77} & 58.59  \ua{\tt\pz 6.78} & 38.47 \ua{\tt\pz 4.67} & 47.42 \ua{\tt\pz 4.17}\\
        \midrule
        \textbf{Mixtral-8x7B} & 73.51 \da{\tt-0.15} & 26.84 \ua{\tt17.60} & 48.99 \ua{\tt\pz 4.33} & 40.66 \da{\tt-3.90} & 27.91 \da{\tt-1.07} & 68.35  \ua{\tt\pz 0.49} & 46.16 \ua{\tt\pz 3.89} & 55.15 \ua{\tt\pz 3.50}\\
        \midrule
        \textbf{Falcon-7B} & 52.40 \da{\tt-2.90} & 25.54 \ua{\tt12.62} & 33.03 \da{\tt-4.42} & 29.18 \da{\tt-3.64} & 20.70 \da{\tt-1.00} & 46.91  \da{\tt-7.71} & 34.91 \da{\tt-0.71} & 36.92 \da{\tt-3.75} \\
        \midrule
        \multicolumn{9}{c}{\textit{Recurrent and Hybrid Models}} \\
        \midrule
        \textbf{RecurrentGemma-2B} & 52.88 \ua{\tt\pz 2.33} & 33.43 \ua{\tt\pz 3.12} & 27.36 \da{\tt-1.85} & 30.66 \da{\tt-1.77} & 18.41 \ua{\tt\pz 2.86} & 42.97  \ua{\tt\pz 3.63} & 36.55 \da{\tt-0.14} & 31.46 \ua{\tt\pz 1.39}\\
        \textbf{RecurrentGemma-9B} & 55.75 \da{\tt-1.67} & 31.67 \ua{\tt12.96} & 36.79 \ua{\tt\pz 2.03} & 36.57 \da{\tt-6.66} & 22.99 \ua{\tt\pz 1.84} & 51.25  \ua{\tt13.36} & 37.61 \da{\tt-0.17} & 42.96 \ua{\tt\pz 3.28}\\
        \midrule
        \textbf{Jamba} & 57.66 \da{\tt-2.90} & 43.36 \da{\tt-5.76} & 39.50 \da{\tt-0.98} & 33.00 \da{\tt-5.88} & 23.72 \da{\tt-15.48} & 59.88  \da{\tt-2.60} & 39.80 \da{\tt-6.91} & 45.95 \da{\tt-0.91} \\
        \midrule
        \textbf{FalconMamba-7B} & 55.80 \ua{\tt\pz 0.73} & 42.97 \da{\tt-1.19} & 39.94 \da{\tt-0.43} & 23.76 \da{\tt-0.85} & 23.89 \da{\tt\pz-0.09} & 61.85  \da{\tt-1.80} & 35.92 \da{\tt-0.26} & 47.02 \da{\tt-0.38}\\
        \bottomrule
        \end{tabular}
    }
    \caption{Changes in performance from the use of instruction-tuning. While factuality does not exhibit a specific trend, faithfulness is shown to improve within attention-based models but not recurrent or hybrid models.}
    \label{tab:instruction_tuning-table}
\end{table*}
% \begin{tcolorbox}[colback=red!5!white,colframe=red!75!black,left*=1mm,right*=1mm,halign=center]
    % \fbox{
\begin{center}
    \textbf{O2: Recurrent models may rely less on memory, leading to more reliance on context.}
\end{center}
    % }
% \end{tcolorbox} 
There are also cases where recurrent models consistently outperform attention equivalents, namely on \textsc{MemoTrap} (right plot in \autoref{fig:o12}). Here, the LLM is prompted to complete a well-known proverb with an ending that deviates from the commonly used ending, testing for over-reliance on knowledge memorized from a training corpus. This suggests that although recurrent models might struggle with long-tail knowledge, they are less prone to ignoring contextual cues not stored within parametric memory, or they more model the dynamics based on the context as opposed to acting as a retrieval system. Noticeable drops exist from \texttt{FalconMamba} (58.8) to \texttt{Falcon} (38.9) and from \texttt{RecurrentGemma} 2B/9B to \texttt{Gemma2} 2B/9B (53.7/43.4 vs 42.7/34.4), suggesting that attention can lead to more hallucinations. It further reveals a benefit of recurrent layers; by not memorizing information directly, there may be a reduced tendency to repeat previous observations~\citep{jelassi2024repeat} and greater focus on the context.

% \begin{tcolorbox}[colback=red!5!white,colframe=red!75!black,left*=2mm,right*=2mm,halign=center]
% \textbf{O3}: Scale is required for emergent qualities.
% \end{tcolorbox}
\begin{center}
    % \fbox{
    \textbf{O3: Scale is required for emergent qualities.}
    % }
\end{center}
However, some scale to the data and model remain important. \texttt{Mamba}, \texttt{Finch} and \texttt{Pythia} fail to draw the same distinctions as the other models, but given the scale of training for these (300B tokens compared to >1T tokens for other models) and the fact that the discussed tasks relate directly to information memorization, it is possible that these phenomena fail to emerge at this data scale. This is particularly evidenced by how these models all perform better with size on \textsc{MemoTrap}, which is designed in a way such that larger models should normally perform worse.

\subsection{On the role of Instruction-Tuning.} The use of instruction-tuning appears to show inconsistent improvements across categories (\autoref{tab:instruction_tuning-table}) and tasks\footnote{See \autoref{fig:instruction_tuning_change-tasks-figure}.}. There exists no strict pattern for overall hallucination; on some tasks, its use is effective for all (ex. \textsc{TruthfulQA}) or no (ex. \textsc{TriviaQA}) models. Smaller trends also exist, such as instruction-following observing the strongest gains.

However, changes in faithfulness are generally {\color{teal} positive} for pure attention models and {\color{magenta} negative} for models with recurrent layers, whereas changes in factuality show no consistent trends. This suggests that instruction-tuning for LLMs may enable particular learning patterns which become irrelevant when using recurrent layers. Their inductive biases therefore can have a particular effect on specific forms of hallucinations emerging, with instruction-tuning being a manner to mitigate these. 

Recalling from \autoref{sec:background}, instruction-tuning can be understood as shifting the projection weights towards a new space through the additional fine-tuning (which is of itself simply an additional step of language modeling). This leads to the over-writing of some of the pre-training corpus with that used for fine-tuning, effectively replacing the keys and values from the underlying retrieval storage. While effects are ambiguous on factuality, as some potentially relevant facts can be removed, faithfulness often improves from this process, as the fine-tuning corpus is more likely to contain information relevant to context following and thereby make it easier for the attention mechanism to retrieve relevant information for such a purpose.
In addition to highlighting how enhancing the ability to follow instructions does not necessarily lead to factually correct results, another conclusion stemming from the fact that recurrent models generally observe no benefits in faithfulness after instruction-tuning is that such models' inductive biases render them inherently more faithful or the process of making them more faithful is distinct from self-attention.

%This pattern suggests that the trade-offs between the two are generally consistent between all different models, namely that the ability to follow instructions closely will not always lead to the produced information being consistently accurate. Hence when using instruction-tuning, ensuring that the instruction is properly being followed might lead the model to produce increasingly inaccurate information due to the desire to adhere more closely to a given context and reduce reliance on internally held knowledge.

\begin{figure}[ht!]
    \centering
    \includegraphics[width=\linewidth]{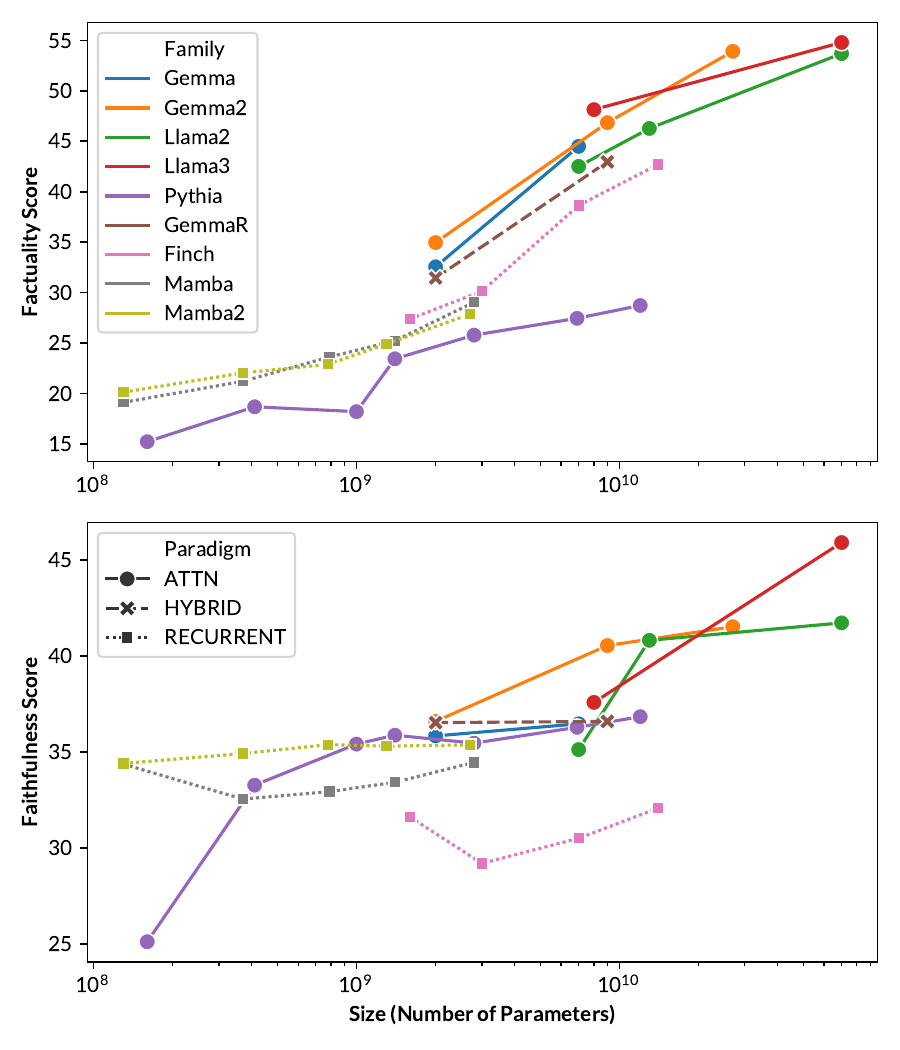}
    \caption{
        Changes in factuality (top) and faithfulness (bottom) as models are increased in size. Score range between 0 and 100. Factuality always increases with the number of parameters, however faithfulness increases are only meaningful for pure-attention models.
    }
    \label{fig:factuality_size}
\end{figure}

\begin{figure}[ht!]
    \centering
    \includegraphics[width=\linewidth]{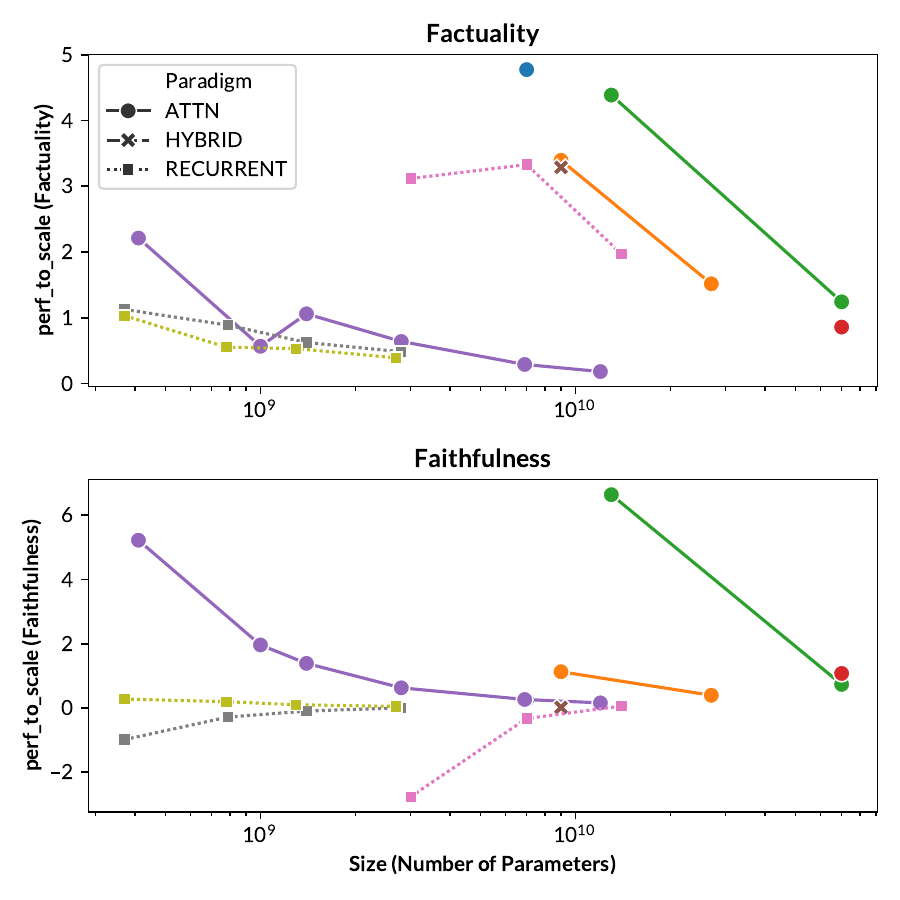}
    \caption{Performance to scale values for factuality and faithfulness. Colors differentiate model families while shapes differentiate the model type. Factuality improves with model size (indicated by values >0) and the model type does not have a link with this relative improvement. Recurrent/hybrid models show low values for faithfulness, indicating that size increases generally do not benefit them, unlike attention models.
    % (\ding{108} for pure attention models, \ding{54} for hybrid models and \ding{110} for pure recurrent models).
    }
    \label{fig:perf_to_scale}
\end{figure}

\subsection{Impact of Model Size on Hallucinations.} With the large amount of prior work~\citep{kaplan2020scalinglawsneurallanguage, wei2022emergent, hoffmann2022trainingcomputeoptimallargelanguage} concluding that model size can play a significant role in model reasoning abilities, we provide this additional axis of variation in \autoref{fig:factuality_size}.

Immediately, we observe that increased size leads to improvements in factuality regardless of model. This is partially expected, considering the range of work that suggest a link between size and the amount of factual knowledge that can be encoded within parametric space. However, faithfulness is different. Recurrent and hybrid architectures tend to saturate at a small size whereas pure attention models do not. For example, \texttt{Mamba} stagnates in faithfulness at around 130M parameter, whereas \texttt{Pythia} continues to increase up to the billions. \texttt{Finch} and \texttt{RecurrentGemma} also exhibit similar behavior. Furthermore, even the smallest \texttt{Mamba} are equally faithful as a 7B \texttt{LLaMA2} and 1.4B \texttt{Pythia}, indicating a meaningful manner in which their hallucinations may differ. To further illustrate, we show a performance to scale metric, defined as
\begin{equation*}
    \texttt{perf\_to\_scale}(m, \mathcal{M}, \mathcal{B}) = \frac{m(\mathcal{M}) - m(\mathcal{B})}{P(\mathcal{M})/P(\mathcal{B}) - 1}
\end{equation*}
where $m(\cdot)$ is a metric of interest, $\mathcal{M}$ is a model and $\mathcal{B}$ is a base model of the family as $\mathcal{M}$. $P(\cdot)$ counts the number of parameters in a given model. Namely, this measures the relative increase in $m$ by linear scaling the model, with a lower value indicating that linearly increasing the number of parameters yields no changes while larger values indicate scaling leads to significant improvements. More positive values indicate that the score increases greatly with linear increases in size and negative values the opposite. Consistently negative or near-zero values hence mean that size has little effect on the score. \autoref{fig:perf_to_scale} depicts the performance to scale changes for factuality and faithfulness, where it becomes apparent that increasing the number of parameters in hybrid and recurrent models yields a significantly lower increase in faithfulness compared to pure attention models.
% This saturation in performance further relates to our prior hypothesis that the hidden state can act as an information bottleneck. Given that few improvements are observed from the transition to larger models, it suggests that more layers and an increased hidden state do not improve this information retention concern, indicating that an the use of a compressed hidden state remains a hurdle for maintaining faithfulness models. While the direct amount of Faithfulness hallucinations that occur is appears to be more related to the quality of the training data (higher quality data leads to fewer issues), the smaller gains remains to be a trait of recurrent architectures in general.

\begin{table*}[ht!]
    \centering
    \resizebox{\linewidth}{!}{
    \begin{tabular}{l|c|cc|cc}
        \toprule
        \multicolumn{1}{c|}{\textbf{Category}} & \textbf{Baseline} (\texttt{LLaMA3-8B}) & \textbf{25\% Reset} & \textbf{50\% Reset} & \textbf{25\% Replaced} & \textbf{50\% Replaced} \\
        \midrule
        Hallucination Detection & 60.76 \ua{\tt10.12} & 50.27 \ua{\tt\pz 3.24}& 44.25 \ua{\tt\pz 8.53}& 53.64 \ua{\tt\pz 0.91}& 53.29 \ua{\tt 0.33}\\
        Instruction Following & 22.06 \ua{\tt26.76} & 14.29 \ua{\tt6.85}& 13.11 \ua{\tt6.50}& 20.81 \da{\tt-0.92}& 20.18 \ua{\tt 0.32}\\
        Closed-Book QA & 41.55 \ua{\tt\pz 1.30} & 36.79 \da{\tt-0.09}& 31.69 \da{\tt-0.07}& 36.26 \da{\tt-0.42}& 33.74 \ua{\tt 0.14}\\
        Reading Comprehension & 33.52 \ua{\tt\pz 3.62} & 30.87 \ua{\tt\pz 1.85} & 29.65 \ua{\tt\pz 2.25} & 32.16 \ua{\tt\pz 0.51} & 30.68 \ua{\tt 0.03} \\
        Summarization & 26.62 \da{\tt-1.60} & 21.12 \ua{\tt\pz 1.06} & 20.86 \ua{\tt\pz 1.67} & 23.84 \ua{\tt\pz 0.54} & 22.59 \ua{\tt 0.79} \\
        Fact-Checking & 60.84 \ua{\tt\pz 4.82} & 54.93 \ua{\tt\pz 2.69} & 50.82 \ua{\tt\pz 3.17} & 55.45 \da{\tt-1.06} & 49.66 \ua{\tt 2.16} \\
        \midrule
        \textsc{Faithfulness} & 37.60 \ua{\tt10.09} & 31.40 \ua{\tt\pz 4.29}& 30.30 \ua{\tt\pz 4.48}& 35.74 \ua{\tt\pz 0.17}& 34.32\ua{\tt 0.13} \\
        \textsc{Factuality} & 48.14 \ua{\tt\pz 1.60} & 43.16 \ua{\tt\pz 0.83}& 38.98 \ua{\tt\pz 1.60}& 42.85 \da{\tt-0.31}& 39.08\ua{\tt 0.56} \\
        \bottomrule
    \end{tabular}
    }
    \caption{Controlling for the pre-training and post-training data as well as the presence of a channel mixer. Baseline is a \texttt{LLaMA3-8B} model. Colored values in brackets represent change in category performance after an additional phase of supervised instruction tuning. Note that the SFT dataset for the baseline differs from other models.}
    \label{tab:distill-table}
\end{table*}

This suggests that recurrent layers enable an inherent ability to follow contexts not improved by model size. Their inductive biases can be perceived as encouraging context following, supported by how the hidden state incorporates information from the context with the current input to control the generation. This perspective aligns these results with our instruction-tuning observations, where such models do not exhibit a clear benefit compared to their attention-based counterparts. In particular, we can interpret this as indicating that the base model learns to perform this output control even without direct instruction-tuning; hence applying these techniques is ineffective as we previously observed.

\subsection{Controlling for Data/Model Differences}

Models are often trained on varying amounts of data, both for pre-training and fine-tuning, while also potentially possessing additional architectural differences in components, such as a channel mixer that exchanges information within each individual token. These can each influence the representation and potentially have a downstream effect on hallucination. To verify whether or not this may be the case, we control for each of these differences in order to quantify the direct effects that exist from replacing attention with sequence layers.

We inspire ourselves from \citet{wang2024the}; we start with a pre-trained base model and replace some attention with sequence layers, namely \texttt{Mamba}, while retaining a similar parameter count and keeping the channel mixing Gated MLPs. For fair comparison, we also consider a scenario where the replaced layers are simply re-initialized. These layers are then trained using a standard next-token prediction objective. Specifically, we use a 8B \texttt{LLaMA3} model and replace/re-initialize 25\% or 50\% of the attention layers in the model. These layers are then trained, providing us with a synthetically pre-trained model for which both the pre-training data and the channel mixer are controlled for, enabling us to directly compare the effects of the use of sequence layers compared to attention layers. As a next step, we also perform an additional end-to-end instruction tuning. Thus we control for:
\begin{enumerate}[itemsep=0ex, leftmargin=0.5cm, label=(\arabic*)]
    \item The pre-training phase, through the base model and dataset used for re-training.
    \item The instruction fine-tuning process, through the use of the same instruction dataset mix.
    \item Additional architectural components through when they are frozen and trained.% As such, the observations made from this experiment stem solely from the changes from a sequence layer to attention as the time/sequence mixing component.
\end{enumerate}
We use a filtered SlimPajama~\citep{cerebras2023slimpajama} to re-train layers with a sequence length of 4096. For the instruction tuning phase, we use the same datasets as \citet{wang2024the}~\citep{genqa, infinityinstruct, openhermes2.5}. We follow hyper-parameters provided by \citet{wang2024the}.%\footnote{\url{https://github.com/jxiw/MambaInLlama/tree/main/mamba_llama}}.

Results in \autoref{tab:distill-table} demonstrate that our previous observations persist. Replacing attention layers with sequence layers and then re-training appears to regain the losses from the initial baseline model, whereas simply resetting the attention layers still appears to observe a noticeable gap. Findings regarding the effects of instruction fine-tuning hold as well, which is that pure-attention models observe a much greater increase in faithfulness relative to factuality, while hybrid models do not.% Overall, this controlled ablation serves as evidence that architecture can directly influence how LLMs perform in terms of hallucination.

\section{Discussion}

\paragraph{Mitigating Hallucinations in LLMs.} %Hallucinations motivate the need to both detect and mitigate them, particularly when they can escalate to have adverse effects. Existing techniques may be insufficient, as they contain potential implicit biases that can fundamentally conflict with the manners in which hallucinations may differ between models. 
Facts such as sequence models saturating in faithfulness in a manner not resolved through instruction-tuning indicates that existing  techniques for hallucination mitigation can differ in their effectiveness due to a bias towards specific architectures. Current techniques therefore deserve a look, in particular in terms of their effectiveness across the different inductive biases of models. While some methods evidently might be specifically catered to specific models, e.g. using attention~\citep{li2023inferencetime, zhang2024-truthx, chuang2024lookbacklensdetectingmitigating}, other methods may implicitly also possess biases due to the underlying assumptions being made or specific requirements that may be unevenly addressed through different architectures, such as the reliance on data-refinement or additional context~\citep{shi2024-trusting}.

\paragraph{Inherent Faithfulness and Factuality.} Designing inherently faithful~\citep{herman2019promiseperilhumanevaluation, wiegreffe-pinter-2019-attention} or factual models remains an important issue. While much work suggests attention-based LLMs lack an inherent tendency to be faithful~\citep{jacovi2020towards, wiegreffe2021teach}, methods have been designed to render models more interpretable~\citep{madsen2024faithfulness} for specifically evaluating this. Similarly, some of these methods have been shown to also adapt to Mamba models~\citep{sharma2024factualmamba}. However, while interpretability is useful for evaluating models as being faithful or factual, it does not ensure an inherent tendency towards being either.

% Nevertheless, these results hint that architectures can provide an avenue through which desiderata can be attained. SSMs present an option which in particular appears suited for settings which require utilizing information provided within a context, given their use within control theory to model system dynamics. With language, this can be analogous to using the fixed-size representation of the previous context for control, which may mean that such models can remain more faithful as the context remains the source that guides the generation. In comparison, attention-based models have been shown to act as retrieval systems that retrieve tokens based on high similarity with memorized information; hence, when a large difference exists between the context and what has been stored from training, breakdowns in faithfulness can occur.

\section{Conclusion}

Are hallucinations the direct result of how token-mixing takes place across time? We study whether different inductive biases of LLM architectures can increase this propensity by observing how hallucinations change across numerous tasks. Patterns emerge demonstrating that some architectural biases can lead to either improvements or degradation in performance compared to others, both for individual tasks and categorizations. Additionally, some model types lead to inherent behaviour when comparing the effects of instruction-tuning and scaling model size. In sum, model-specific inductive biases can have a direct effect on the type of hallucinations they are faced with. However, the data used and the specific model construction can also affect learning. We hope that future work can build on our findings by exploring both the types of hallucinations that can occur and techniques meant to mitigate them and how to design more universal techniques to make models more robust or reliable.

\section{Limitations}

\paragraph{Evaluating faithfulness and factuality.} A major limitation existing in hallucination detection and mitigation research is the lack of metrics that provide explicit information regarding hallucination. While the development of such metrics is an active area of research, limitations still exist, such as biases when involving additional models within the evaluation process.

\paragraph{Limits on tasks.} Another limitation of this work is the non-exhaustive set of tasks and domains in which we measure hallucinations. Due to the exhaustive ways in which hallucinations can be measured, we limit ourselves to tasks that are well motivated and frequently used in practice.

\section{Ethical Concerns}

This paper provides an analysis on the effects of different architectural inductive biases on the propensity to hallucinate within large language models. As such, mistakes in methodology can lead to unsupported confidence or skepticism regarding their performance with the explored task or related ones. While skepticism may not be an ethical issue, unsupported confidence can be problematic. However, the overall message is that all LLMs have specific tasks and settings in which they will display a greater propensity to hallucinate, hence we do not believe that the ideas expressed in this work will explicitly lead to unsupported confidence.

\section{Acknowledgements}

Jerry Huang was supported by a National Science and Engineering Research Council (NSERC) Canada Graduate Scholarship, a Fonds de Recherche du Qu\'{e}bec Nature et technologies (FRQNT) Training Scholarship and a Hydro-Qu\'{e}bec Excellence Scholarship. Sarath Chandar is supported by a Canada CIFAR AI Chair, the Canada Research Chair in Lifelong Machine Learning and a NSERC Discovery Grant. % The experiments were in part enabled by computational resources provided by Calcul Québec (\url{calculquebec.ca}), the Digital Research Alliance of Canada (\url{alliancecan.ca}) and NVIDIA.

\bibliography{refs}
\appendix
\section{Technical Implementation Details}\label{app:technical_details}

\subsection{Models and Datasets}

All models and datasets used in this paper are public and directly available on the HuggingFace Hub.

\subsection{Experimental Setup}\label{app:experimental_setup}

Experiments were run using the Language Model Evaluation Harness~\citep{eval-harness}.

\subsection{Datasets} We evaluate on the Hallucination Leaderboard~\citep{hong2024hallucinationsleaderboardopen}, consisting of tasks
\begin{enumerate}[itemsep=0ex, leftmargin=0.5cm, label=\arabic*)]
    \item \textbf{Closed-book Open-domain QA}$^{\color{magenta}\clubsuit}$: 
        \textsc{NQ-Open}~\citep{kwiatkowski2019-natural}, 
        \textsc{TriviaQA}~\citep{joshi2017-triviaqa}, 
        \textsc{TruthfulQA} (MC1, MC2, Generative)~\citep{lin2022truthfulqa}, 
        \textsc{PopQA}~\citep{mallen2023trust}
    \item \textbf{Summarization}$^\spadesuit$: 
        \textsc{XSum}~\citep{Narayan2018DontGM},    
        \textsc{CNN/DM}~\citep{see2017-get}
    \item \textbf{Reading Comprehension}$^\spadesuit$: 
        \textsc{RACE}~\citep{lai2017-race}, 
        \textsc{SQuADv2}~\citep{rajpurkar2018-know}, 
        \textsc{NQ-Swap}~\citep{longpre2021-entity}
    \item \textbf{Instruction Following}$^\spadesuit$: 
        \textsc{MemoTrap}~\citep{liu2023memotrap}, 
        \textsc{IFEval}~\citep{zhou2023instructionfollowing}
    \item \textbf{Hallucination Detection}: 
        \textsc{FaithDial}$^\spadesuit$~\citep{dziri2022faithdial}, 
        \textsc{HaluEval}$^\spadesuit$ (QA, Summarization, Dialogue)~\citep{li2023halueval}, 
        \textsc{True-False}$^{\color{magenta}\clubsuit}$~\citep{azaria-mitchell-2023internal}
    \item \textbf{Fact Checking}$^{\color{magenta}\clubsuit}$: 
        \textsc{FEVER}~\citep{thorne2018-fever}
\end{enumerate}
% This enables us to evaluate effects along different axes, in particular within specific datasets, domains or even related concepts. Each evaluation uses different evaluation metrics listed in their respective results columns. 
For tasks, a higher score (ranging from 0 to 100) indicates better performance. These are
\begin{itemize}
    \item EM: \textsc{HaluEval} (all sets), \textsc{PopQA}, \textsc{TruthfulQA} (MC1 and MC2), \textsc{SQuADv2}, \textsc{NaturalQuestions} (\textsc{NQ-Open}), \textsc{TriviaQA}
    \item Accuracy: \textsc{MemoTrap}, \textsc{TruthfulQA} (Gen), \textsc{RACE}, \textsc{IfEval}
    \item Rouge-L: \textsc{CNNDM}, \textsc{XSum}, \textsc{FaithDial}, \textsc{TrueFalse}, \textsc{Fever10}
\end{itemize}

Tasks are further divided into two categories: \textit{faithfulness} ($\spadesuit$) hallucinations, i.e. whether an LLM generation adheres to the given source of information, and \textit{factuality} (${\color{magenta}\clubsuit}$) hallucinations, i.e. whether LLMs generate factually correct content according to world knowledge based on knowledge acquired during training. To compute scores across such categories, the scores for each task in the category are averaged.

\subsection{Models and Baselines} 
For our comparison, we use various models (\autoref{tab:models}) that vary in size and architecture, with a particular focus on different time-mixing methods.
These include:
\begin{itemize}
    \item All Pythia models
    \item All publicly available Mamba~\citep{gu2024mamba, dao2024transformers} models
    \item All LLaMA2 and LLaMA3~\citep{touvron2023llama} models
    \item Falcon-7B~\citep{almazrouei2023falconseriesopenlanguage} and FalconMamba-7B
    \item Mistral-7B~\citep{jiang2023mistral}
    \item All Gemma~\citep{gemma_2024} models
    \item RWKV~\citep{peng2023rwkv} models, namely Finch (RWKV-v6)
    \item RecurrentGemma~\citep{botev2024recurrentgemma}
    \item Mixtral-8x7B~\citep{jiang2024mixtralexperts}
    \item Jamba~\citep{jamba}
\end{itemize} 
For models with instruction fine-tuned versions, we use both the base version as well as the instruction fine-tuned variant.

\begin{table*}[ht!]
    \resizebox{0.98\linewidth}{!}{
        \centering
        \begin{tabular}{llc}
        \toprule
        Model & Public Link & HuggingFace Model \\
        \midrule
        \texttt{Pythia-160M} &  \href{https://huggingface.co/EleutherAI/pythia-160m}{\tt EleutherAI/pythia-160m} & \ding{52} \\
        \texttt{Pythia-410M} &  \href{https://huggingface.co/EleutherAI/pythia-410m}{\tt EleutherAI/pythia-410m} & \ding{52} \\
        \texttt{Pythia-1B} &  \href{https://huggingface.co/EleutherAI/pythia-1b}{\tt EleutherAI/pythia-1b} & \ding{52} \\
        \texttt{Pythia-1.4B} &  \href{https://huggingface.co/EleutherAI/pythia-1.4b}{\tt EleutherAI/pythia-1.4b} & \ding{52} \\
        \texttt{Pythia-2.8B} &  \href{https://huggingface.co/EleutherAI/pythia-2.8b}{\tt EleutherAI/pythia-2.8b} & \ding{52} \\
        \texttt{Pythia-6.9B} &  \href{https://huggingface.co/EleutherAI/pythia-6.9b}{\tt EleutherAI/pythia-6.9b} & \ding{52} \\
        \texttt{Pythia-12B} &  \href{https://huggingface.co/EleutherAI/pythia-12b}{\tt EleutherAI/pythia-12b} & \ding{52} \\
        \midrule
        \texttt{Mamba-130M} &  \href{https://huggingface.co/state-spaces/mamba-130m}{\tt state-spaces/mamba-130m} & \ding{56} \\
        \texttt{Mamba-370M} &  \href{https://huggingface.co/state-spaces/mamba-370m}{\tt state-spaces/mamba-370m} & \ding{56} \\
        \texttt{Mamba-790M} &  \href{https://huggingface.co/state-spaces/mamba-790m}{\tt state-spaces/mamba-790m} & \ding{56} \\
        \texttt{Mamba-1.4B} &  \href{https://huggingface.co/state-spaces/mamba-1.4b}{\tt state-spaces/mamba-1.4b} & \ding{56} \\
        \texttt{Mamba-2.8B} &  \href{https://huggingface.co/state-spaces/mamba-2.8m}{\tt state-spaces/mamba-2.8b} & \ding{56} \\
        \texttt{Mamba2-130M} &  \href{https://huggingface.co/state-spaces/mamba2-130m}{\tt state-spaces/mamba2-130m} & \ding{56} \\
        \texttt{Mamba2-370M} &  \href{https://huggingface.co/state-spaces/mamba2-370m}{\tt state-spaces/mamba2-370m} & \ding{56} \\
        \texttt{Mamba2-780M} &  \href{https://huggingface.co/state-spaces/mamba2-780m}{\tt state-spaces/mamba2-780m} & \ding{56} \\
        \texttt{Mamba2-1.3B} &  \href{https://huggingface.co/state-spaces/mamba2-1.3b}{\tt state-spaces/mamba2-1.3b} & \ding{56} \\
        \texttt{Mamba2-2.7B} &  \href{https://huggingface.co/state-spaces/mamba2-2.7b}{\tt state-spaces/mamba2-2.7b} & \ding{56} \\
        \texttt{Mamba2Attention-2.7B} & \href{https://huggingface.co/state-spaces/mamba2attn-2.7b}{\tt state-spaces/mamba2attn-2.7b} & \ding{56}\\
        \midrule
        \texttt{RecurrentGemma-2B} & \href{https://huggingface.co/google/recurrentgemma-2b}{\tt google/recurrentgemma-2b} & \ding{52} \\
        \texttt{RecurrentGemma-2B (IT)} & \href{https://huggingface.co/google/recurrentgemma-2b-it}{\tt google/recurrentgemma-2b-it} & \ding{52} \\
        \texttt{RecurrentGemma-9B} & \href{https://huggingface.co/google/recurrentgemma-9b}{\tt google/recurrentgemma-9b} & \ding{52} \\
        \texttt{RecurrentGemma-9B (IT)} & \href{https://huggingface.co/google/recurrentgemma-9b-it}{\tt google/recurrentgemma-9b-it} & \ding{52} \\
        \midrule
        \texttt{Gemma-2B} & \href{https://huggingface.co/google/gemma-2b}{\tt google/gemma-2b} & \ding{52} \\
        \texttt{Gemma-2B (IT)} & \href{https://huggingface.co/google/gemma-2b-it}{\tt google/gemma-2b-it} & \ding{52} \\
        \texttt{Gemma-9B} & \href{https://huggingface.co/google/gemma-9b}{\tt google/gemma-9b} & \ding{52} \\
        \texttt{Gemma-9B (IT)} & \href{https://huggingface.co/google/gemma-9b-it}{\tt google/gemma-9b-it} & \ding{52} \\
        \midrule
        \texttt{Gemma2-2B} & \href{https://huggingface.co/google/gemma2-2b}{\tt google/gemma2-2b} & \ding{52} \\
        \texttt{Gemma2-2B (IT)} & \href{https://huggingface.co/google/gemma2-2b-it}{\tt google/gemma2-2b-it} & \ding{52} \\
        \texttt{Gemma2-9B} & \href{https://huggingface.co/google/gemma2-9b}{\tt google/gemma2-9b} & \ding{52} \\
        \texttt{Gemma2-9B (IT)} & \href{https://huggingface.co/google/gemma2-9b-it}{\tt google/gemma2-9b-it} & \ding{52} \\
        \texttt{Gemma2-27B} & \href{https://huggingface.co/google/gemma2-27b}{\tt google/gemma2-27b} & \ding{52} \\
        \texttt{Gemma2-27B (IT)} & \href{https://huggingface.co/google/gemma2-27b-it}{\tt google/gemma2-27b-it} & \ding{52} \\
        \midrule
        \texttt{Falcon-7B} & \href{https://huggingface.co/tiiuae/falcon-7b}{\tt tiiuae/falcon-7b} & \ding{52} \\
        \texttt{Falcon-7B (IT)} & \href{https://huggingface.co/tiiuae/tiiuae/falcon-7b-instruct}{\tt tiiuae/falcon-7b-instruct} & \ding{52} \\
        \midrule
        \texttt{FalconMamba-7B} & \href{https://huggingface.co/tiiuae/falcon-mamba-7b}{\tt tiiuae/falcon-mamba-7b} & \ding{52} \\
        \texttt{FalconMamba-7B (IT)} & \href{https://huggingface.co/tiiuae/tiiuae/falcon-mamba-7b-instruct}{\tt tiiuae/falcon-mamba-7b-instruct} & \ding{52} \\
        \midrule
        \texttt{Mistral-7B} & \href{https://huggingface.co/mistralai/Mistral-7B-v0.3}{\tt mistralai/Mistral-7B-v0.3} & \ding{52} \\
        \texttt{Mistral-7B (IT)} & \href{https://huggingface.co/mistralai/Mistral-7B-Instruct-v0.3}{\tt mistralai/Mistral-7B-Instruct-v0.3} & \ding{52} \\
        \midrule
        \texttt{Mixtral-8x7B} & \href{https://huggingface.co/mistralai/Mixtral-8x7B-v0.1}{\tt mistralai/Mixtral-8x7B-v0.1} & \ding{52} \\
        \texttt{Mixtral-8x7B (IT)} & \href{https://huggingface.co/mistralai/Mixtral-8x7B-Instruct-v0.1}{\tt  mistralai/Mixtral-8x7B-Instruct-v0.1} & \ding{52} \\
        \midrule
        \texttt{Jamba} & \href{https://huggingface.co/ai21labs/Jamba-v0.1}{\tt ai21labs/Jamba-v0.1} & \ding{52} \\
        \texttt{Jamba (IT)} & \href{https://huggingface.co/ai21labs/AI21-Jamba-1.5-Mini}{\tt ai21labs/AI21-Jamba-1.5-Mini} & \ding{52} \\
        \midrule
        \texttt{LLaMA2-7B} & \href{https://huggingface.co/meta-llama/Llama-2-7b-hf}{\tt meta-llama/Llama-2-7b-hf} & \ding{52} \\
        \texttt{LLaMA2-7B (IT)} & \href{https://huggingface.co/meta-llama/Llama-2-7b-chat-hf}{\tt meta-llama/Llama-2-7b-hf} & \ding{52} \\
        \texttt{LLaMA2-13B} & \href{https://huggingface.co/meta-llama/Llama-2-13b-hf}{\tt meta-llama/Llama-2-13b-hf} & \ding{52} \\
        \texttt{LLaMA2-13B (IT)} & \href{https://huggingface.co/meta-llama/Llama-2-13b-chat-hf}{\tt meta-llama/Llama-2-13b-hf} & \ding{52} \\
        \texttt{LLaMA2-70B} & \href{https://huggingface.co/meta-llama/Llama-2-70b-hf}{\tt meta-llama/Llama-2-70b-hf} & \ding{52} \\
        \texttt{LLaMA2-70B (IT)} & \href{https://huggingface.co/meta-llama/Llama-2-70b-chat-hf}{\tt meta-llama/Llama-2-70b-hf} & \ding{52} \\
        \midrule
        \texttt{LLaMA3-8B} & \href{https://huggingface.co/meta-llama/Meta-Llama-3-8B}{\tt https://huggingface.co/meta-llama/Meta-Llama-3-8B} & \ding{52} \\
        \texttt{LLaMA3-8B (IT)} & \href{https://huggingface.co/meta-llama/Meta-Llama-3-8B-Instruct}{\tt https://huggingface.co/meta-llama/Meta-Llama-3-8B-Instruct} & \ding{52} \\
        \texttt{LLaMA3-70B} & \href{https://huggingface.co/meta-llama/Meta-Llama-3-70B}{\tt https://huggingface.co/meta-llama/Meta-Llama-3-70B} & \ding{52} \\
        \texttt{LLaMA3-70B (IT)} & \href{https://huggingface.co/meta-llama/Meta-Llama-3-70B-Instruct}{\tt https://huggingface.co/meta-llama/Meta-Llama-3-70B-Instruct} & \ding{52} \\
        \bottomrule
        \end{tabular}
    }
    \caption{Models used and public links to their weights.}
    \label{tab:models}
\end{table*}

\subsection{Computing Resources Used}

All results were obtained using a server of 8 NVIDIA V100 32GB or 4 NVIDIA RTX A6000 48GB GPUs. The \texttt{accelerate} package was used for model sharding in instances where a single GPU was insufficient to store the entire model.

\section{Complete Results}\label{app:complete_tables}

% \begin{figure*}[ht!]
%     \centering
%     \includegraphics[width=\linewidth]{figures/tasks.pdf}
%     \caption{Complete results of all model performance on all tasks.}
%     \label{fig:tasks_heatmap}
% \end{figure*}

\begin{figure*}[ht!]
    \centering
    \caption{Performance of various base models on tasks within the Hallucination Leaderboard.}
    \includegraphics[width=0.95\linewidth]{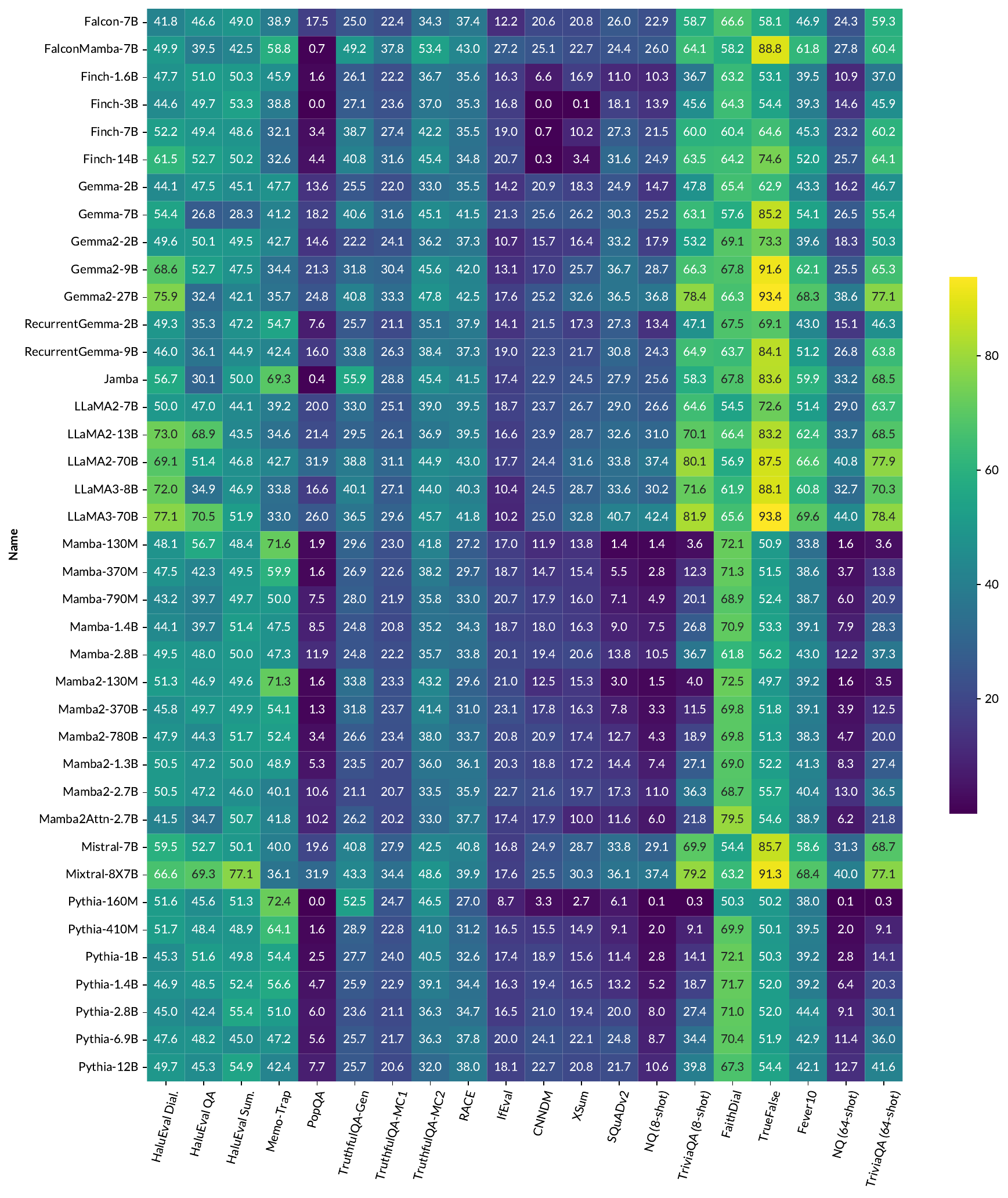}
    \label{fig:tasks-base-figure}
    \vspace{-0.25cm}
\end{figure*}

\begin{table*}[ht!]
    \centering
    \resizebox{\linewidth}{!}{
    % \begin{tabular}{lrrrrrrrrrrrrrrrrrrrr|c}
        \begin{tabular}{lccccccccccccccccccccc}
            \toprule
            \multicolumn{1}{c}{\multirow{2}{*}{\textbf{Model}}} & \textsc{HaluEval Dial.} & \textsc{HaluEval QA} & \textsc{HaluEval Sum.} & \textsc{Memo-Trap} & \textsc{PopQA} & \textsc{TruthfulQA-Gen} & \textsc{TruthfulQA-MC1} & \textsc{TruthfulQA-MC2} & \textsc{RACE} & \textsc{IfEval} & \textsc{CNNDM} & \textsc{XSum} & \textsc{SQuADv2} & \textsc{NQ (8-shot)} & \textsc{TriviaQA (8-shot)} & \textsc{FaithDial} & \textsc{TrueFalse} & \textsc{Fever10} & \textsc{NQ (64-shot)} & \textsc{TriviaQA (64-shot)} & \multirow{2}{*}{\textbf{Avg}} \\
            & EM & EM & EM & Accuracy & EM & Rouge-L & Accuracy & Accuracy & Accuracy & Accuracy & Rouge-L & Rouge-L & EM & EM & EM & Rouge-L & Rouge-L & Rouge-L & EM & EM & \\
            \midrule
            \textbf{Falcon-7B} & 41.81 & 46.57 & 48.95 & 38.89 & 17.46 & 24.97 & 22.40 & 34.26 & 37.42 & 12.20 & 20.59 & 20.81 & 26.01 & 22.88 & 58.68 & 66.60 & 58.08 & 46.91 & 24.32 & 59.27 & 35.87 \\
            \textbf{Falcon-7B (IT)} & 38.40 & 29.64 & 48.65 & 60.79 & 8.68 & 38.07 & 29.01 & 44.12 & 37.22 & 15.53 & 20.43 & 18.97 & 20.68 & 14.29 & 39.21 & 67.14 & 63.66 & 39.20 & 15.51 & 39.95 & 33.71 \\
            \midrule
            \textbf{FalconMamba-7B} & 49.93 & 39.54 & 42.53 & 58.76 & 0.74 & 49.20 & 37.82 & 53.40 & 42.97 & 27.17 & 25.09 & 22.68 & 24.36 & 25.96 & 64.13 & 58.18 & 88.84 & 61.85 & 27.84 & 60.44 & 41.21 \\
            \textbf{FalconMamba-7B (IT)} & 50.14 & 40.16 & 39.41 & 56.20 & 0.18 & 49.08 & 37.70 & 53.23 & 42.58 & 27.36 & 24.79 & 22.80 & 24.53 & 26.34 & 63.45 & 62.67 & 90.29 & 60.05 & 27.65 & 58.45 & 40.89 \\
            \midrule
            \textbf{Finch-1.6B} & 47.72 & 51.05 & 50.32 & 45.94 & 1.58 & 26.07 & 22.15 & 36.73 & 35.60 & 16.27 & 6.63 & 16.91 & 10.97 & 10.30 & 36.71 & 63.24 & 53.10 & 39.47 & 10.91 & 37.05 & 29.62 \\
            \textbf{Finch-3B} & 44.65 & 49.71 & 53.34 & 38.78 & 0.05 & 27.05 & 23.62 & 37.02 & 35.31 & 16.82 & 0.02 & 0.05 & 18.13 & 13.85 & 45.58 & 64.28 & 54.45 & 39.26 & 14.60 & 45.89 & 29.65 \\
            \textbf{Finch-7B} & 52.17 & 49.41 & 48.60 & 32.05 & 3.45 & 38.68 & 27.42 & 42.19 & 35.50 & 19.04 & 0.66 & 10.19 & 27.33 & 21.50 & 60.03 & 60.44 & 64.57 & 45.26 & 23.19 & 60.19 & 34.39 \\
            \textbf{Finch-14B} & 61.52 & 52.74 & 50.18 & 32.59 & 4.39 & 40.76 & 31.58 & 45.43 & 34.83 & 20.70 & 0.33 & 3.45 & 31.60 & 24.93 & 63.48 & 64.23 & 74.58 & 51.98 & 25.65 & 64.12 & 37.13 \\
            \midrule
            \textbf{Gemma-2B} & 44.13 & 47.54 & 45.09 & 47.72 & 13.58 & 25.46 & 22.03 & 33.03 & 35.46 & 14.16 & 20.93 & 18.34 & 24.91 & 14.65 & 47.76 & 65.43 & 62.88 & 43.28 & 16.20 & 46.72 & 34.81 \\
            \textbf{Gemma-7B} & 54.37 & 26.78 & 28.30 & 41.17 & 18.20 & 40.64 & 31.58 & 45.09 & 41.48 & 21.33 & 25.65 & 26.22 & 30.34 & 25.21 & 63.07 & 57.60 & 85.23 & 54.14 & 26.45 & 55.36 & 39.77 \\
            \textbf{Gemma-2B (IT)} & 50.82 & 51.90 & 51.82 & 58.94 & 2.84 & 38.68 & 29.01 & 45.85 & 41.52 & 22.52 & 28.61 & 18.74 & 29.97 & 8.14 & 26.48 & 38.62 & 55.55 & 38.06 & 8.67 & 26.69 & 31.55 \\
            \textbf{Gemma-7B (IT)} & 65.04 & 61.48 & 47.03 & 56.20 & 12.03 & 59.00 & 29.13 & 46.88 & 42.01 & 31.98 & 20.91 & 19.89 & 28.98 & 12.11 & 39.10 & 53.60 & 78.65 & 55.74 & 12.83 & 39.14 & 38.84 \\
            \textbf{Gemma1.1-2B (IT)} & 49.87 & 49.48 & 52.16 & 56.77 & 4.68 & 47.98 & 29.01 & 45.46 & 41.19 & 25.77 & 24.31 & 17.69 & 30.93 & 9.09 & 28.74 & 38.38 & 63.14 & 34.65 & 9.31 & 29.17 & 32.99 \\
            \textbf{Gemma1.1-7B (IT)} & 64.33 & 27.11 & 38.18 & 48.82 & 9.32 & 61.69 & 34.15 & 51.05 & 41.91 & 44.18 & 22.45 & 22.20 & 33.78 & 16.26 & 51.42 & 61.03 & 85.70 & 58.96 & 17.34 & 52.03 & 40.29 \\
            \midrule
            \textbf{Gemma2-2B} & 49.57 & 50.14 & 49.53 & 42.74 & 14.64 & 22.15 & 24.11 & 36.23 & 37.32 & 10.72 & 15.66 & 16.43 & 33.18 & 17.87 & 53.16 & 69.09 & 73.33 & 39.57 & 18.31 & 50.28 & 35.83 \\
            \textbf{Gemma2-9B} & 68.56 & 52.74 & 47.54 & 34.40 & 21.35 & 31.82 & 30.35 & 45.57 & 42.01 & 13.12 & 17.02 & 25.73 & 36.65 & 28.67 & 66.26 & 67.82 & 91.55 & 62.06 & 25.48 & 65.34 & 43.55 \\
            \textbf{Gemma2-27B} & 75.94 & 32.38 & 42.14 & 35.68 & 24.81 & 40.76 & 33.29 & 47.81 & 42.49 & 17.56 & 25.24 & 32.61 & 36.52 & 36.84 & 78.42 & 66.29 & 93.38 & 68.28 & 38.56 & 77.05 & 47.44 \\
            \textbf{Gemma2-2B (IT)} & 59.45 & 34.61 & 46.97 & 48.82 & 11.43 & 43.21 & 36.96 & 53.15 & 44.78 & 26.43 & 15.05 & 16.10 & 37.71 & 17.12 & 49.70 & 68.30 & 85.51 & 58.53 & 17.89 & 45.90 & 38.96 \\
            \textbf{Gemma2-9B (IT)} & 73.18 & 60.80 & 69.62 & 54.06 & 16.80 & 41.74 & 42.84 & 60.11 & 46.99 & 53.60 & 16.64 & 19.41 & 40.05 & 25.18 & 63.93 & 64.76 & 91.41 & 68.53 & 23.27 & 66.04 & 49.00 \\
            \textbf{Gemma2-27B (IT)} & 79.85 & 65.31 & 73.38 & 48.50 & 20.28 & 43.94 & 46.39 & 64.48 & 50.14 & 71.90 & 26.02 & 30.35 & 40.23 & 34.32 & 77.52 & 60.67 & 93.11 & 69.28 & 36.23 & 77.22 & 55.38 \\
            \midrule
            \textbf{RecurrentGemma-2B} & 49.28 & 35.29 & 47.21 & 54.74 & 7.64 & 25.70 & 21.05 & 35.10 & 37.94 & 14.12 & 21.50 & 17.32 & 27.29 & 13.41 & 47.11 & 67.54 & 69.06 & 42.97 & 15.10 & 46.34 & 33.26 \\
            \textbf{RecurrentGemma-9B} & 45.97 & 36.14 & 44.85 & 42.38 & 16.01 & 33.78 & 26.32 & 38.37 & 37.28 & 18.96 & 22.25 & 21.73 & 30.76 & 24.35 & 64.86 & 63.68 & 84.11 & 51.25 & 26.84 & 63.77 & 40.16 \\
            \textbf{RecurrentGemma-2B (IT)} & 46.69 & 35.76 & 48.57 & 51.28 & 5.75 & 33.78 & 26.44 & 42.77 & 39.62 & 21.81 & 22.65 & 19.88 & 30.05 & 11.14 & 36.24 & 67.22 & 77.81 & 46.60 & 11.55 & 36.37 & 34.71 \\
            \textbf{RecurrentGemma-9B (IT)} & 60.28 & 18.42 & 44.47 & 54.98 & 18.77 & 45.53 & 32.19 & 47.60 & 43.31 & 32.27 & 23.62 & 24.05 & 34.12 & 21.86 & 60.37 & 55.99 & 87.25 & 64.61 & 23.52 & 60.69 & 41.63 \\
            \midrule
            \textbf{LLaMA2-7B} & 49.96 & 47.01 & 44.12 & 39.21 & 20.01 & 33.05 & 25.09 & 38.99 & 39.52 & 18.67 & 23.66 & 26.73 & 29.00 & 26.65 & 64.64 & 54.45 & 72.59 & 51.38 & 28.98 & 63.75 & 38.65 \\
            \textbf{LLaMA2-13B} & 73.00 & 68.88 & 43.50 & 34.58 & 21.43 & 29.50 & 26.07 & 36.90 & 39.48 & 16.56 & 23.92 & 28.74 & 32.61 & 31.00 & 70.08 & 66.45 & 83.17 & 62.35 & 33.68 & 68.54 & 43.95 \\
            \textbf{LLaMA2-70B} & 69.12 & 51.44 & 46.76 & 42.74 & 31.90 & 38.80 & 31.09 & 44.86 & 42.97 & 17.74 & 24.38 & 31.63 & 33.77 & 37.37 & 80.07 & 56.91 & 87.49 & 66.63 & 40.83 & 77.88 & 47.43 \\
            \textbf{LLaMA2-7B (IT)} & 58.17 & 49.06 & 45.51 & 50.53 & 7.03 & 42.96 & 29.87 & 44.62 & 44.02 & 25.32 & 27.80 & 23.97 & 33.01 & 23.63 & 57.40 & 56.34 & 80.56 & 57.26 & 24.74 & 55.80 & 40.69 \\
            \textbf{LLaMA2-13B (IT)} & 67.62 & 58.38 & 48.70 & 42.35 & 9.63 & 42.11 & 28.03 & 43.96 & 47.12 & 32.24 & 27.60 & 28.60 & 35.97 & 27.01 & 66.44 & 64.75 & 85.16 & 60.71 & 29.22 & 65.57 & 43.13 \\
            \textbf{LLaMA2-70B (IT)} & 97.26 & 63.49 & 60.52 & 55.88 & 12.95 & 48.35 & 35.74 & 52.75 & 43.92 & 38.82 & 24.73 & 30.17 & 35.75 & 32.22 & 73.76 & 58.52 & 85.83 & 65.18 & 35.29 & 71.47 & 49.37 \\
            \midrule
            \textbf{LLaMA3-8B} & 72.04 & 34.85 & 46.92 & 33.76 & 16.56 & 40.15 & 27.05 & 43.96 & 40.29 & 10.35 & 24.51 & 28.72 & 33.61 & 30.19 & 71.59 & 61.88 & 88.12 & 60.84 & 32.69 & 70.26 & 42.62 \\
            \textbf{LLaMA3-70B} & 77.09 & 70.55 & 51.85 & 33.01 & 26.02 & 36.47 & 29.62 & 45.72 & 41.82 & 10.17 & 24.98 & 32.84 & 40.70 & 42.41 & 81.87 & 65.56 & 93.84 & 69.57 & 44.04 & 78.39 & 50.15 \\
            \textbf{LLaMA3-8B (IT)} & 76.01 & 68.15 & 64.35 & 59.19 & 15.81 & 47.86 & 35.99 & 51.60 & 46.03 & 38.45 & 24.56 & 25.49 & 37.41 & 27.51 & 67.56 & 56.97 & 88.94 & 65.66 & 28.98 & 67.49 & 48.66 \\
            \textbf{LLaMA3-70B (IT)} & 82.79 & 88.26 & 72.77 & 41.88 & 27.67 & 50.31 & 44.06 & 61.83 & 46.99 & 52.50 & 24.40 & 30.06 & 40.23 & 36.45 & 80.20 & 65.02 & 93.53 & 70.56 & 38.56 & 72.89 & 54.75 \\
            \midrule
            \textbf{Jamba} & 56.70 & 30.10 & 50.05 & 69.34 & 0.45 & 55.94 & 28.76 & 45.39 & 41.53 & 17.38 & 22.93 & 24.52 & 27.89 & 25.57 & 58.29 & 67.82 & 83.63 & 59.88 & 33.16 & 68.47 & 42.73 \\
            \textbf{Jamba (IT)} & 55.11 & 20.45 & 49.32 & 74.47 & 0.00 & 64.99 & 32.93 & 49.42 & 44.98 & 0.74 & 7.34 & 9.13 & 32.20 & 19.75 & 50.98 & 63.94 & 85.00 & 57.28 & 27.51 & 62.57 & 38.68 \\
            \midrule
            \textbf{Mamba-130M} & 48.08 & 56.71 & 48.36 & 71.58 & 1.93 & 29.62 & 23.01 & 41.75 & 27.18 & 17.01 & 11.87 & 13.81 & 1.43 & 1.44 & 3.58 & 72.14 & 50.91 & 33.84 & 1.58 & 3.59 & 27.11 \\
            \textbf{Mamba-370M} & 47.53 & 42.29 & 49.48 & 59.94 & 1.60 & 26.93 & 22.64 & 38.22 & 29.67 & 18.67 & 14.68 & 15.41 & 5.49 & 2.80 & 12.33 & 71.29 & 51.49 & 38.55 & 3.71 & 13.80 & 27.15 \\
            \textbf{Mamba-790M} & 43.22 & 39.69 & 49.74 & 50.00 & 7.54 & 28.03 & 21.91 & 35.78 & 33.01 & 20.70 & 17.88 & 16.04 & 7.09 & 4.93 & 20.08 & 68.95 & 52.44 & 38.73 & 5.96 & 20.88 & 28.51 \\
            \textbf{Mamba-1.4B} & 44.07 & 39.72 & 51.36 & 47.54 & 8.51 & 24.85 & 20.81 & 35.21 & 34.26 & 18.67 & 17.95 & 16.28 & 8.95 & 7.48 & 26.84 & 70.90 & 53.34 & 39.08 & 7.89 & 28.25 & 29.53 \\
            \textbf{Mamba-2.8B} & 49.54 & 48.03 & 49.96 & 47.33 & 11.90 & 24.85 & 22.15 & 35.71 & 33.78 & 20.15 & 19.41 & 20.60 & 13.80 & 10.50 & 36.70 & 61.77 & 56.19 & 42.99 & 12.19 & 37.30 & 31.89 \\
            \midrule
            \textbf{Mamba2-130M} & 51.32 & 46.95 & 49.60 & 71.30 & 1.62 & 33.78 & 23.26 & 43.17 & 29.61 & 20.96 & 12.49 & 15.27 & 3.03 & 1.50 & 4.03 & 72.52 & 49.75 & 39.17 & 1.55 & 3.53 & 27.09 \\
            \textbf{Mamba2-370B} & 45.84 & 49.71 & 49.89 & 54.06 & 1.30 & 31.82 & 23.75 & 41.44 & 31.00 & 23.11 & 17.80 & 16.30 & 7.82 & 3.30 & 11.46 & 69.79 & 51.83 & 39.08 & 3.93 & 12.51 & 28.79 \\
            \textbf{Mamba2-780B} & 47.88 & 44.28 & 51.65 & 52.39 & 3.44 & 26.56 & 23.38 & 38.02 & 33.73 & 20.78 & 20.93 & 17.36 & 12.70 & 4.35 & 18.87 & 69.78 & 51.27 & 38.35 & 4.74 & 19.98 & 28.92 \\
            \textbf{Mamba2-1.3B} & 50.48 & 47.19 & 49.97 & 48.93 & 5.26 & 23.50 & 20.69 & 35.98 & 36.08 & 20.33 & 18.78 & 17.19 & 14.42 & 7.37 & 27.12 & 69.00 & 52.23 & 41.26 & 8.34 & 27.39 & 30.36 \\
            \textbf{Mamba2-2.7B} & 50.49 & 47.23 & 46.04 & 40.06 & 10.56 & 21.05 & 20.69 & 33.50 & 35.89 & 22.74 & 21.59 & 19.73 & 17.26 & 11.02 & 36.27 & 68.69 & 55.66 & 40.35 & 12.99 & 36.47 & 32.33 \\
            \textbf{Mamba2Attn-2.7B} & 41.48 & 34.69 & 50.71 & 41.77 & 10.16 & 26.19 & 20.20 & 32.95 & 37.70 & 17.38 & 17.91 & 10.02 & 11.62 & 6.01 & 21.76 & 79.51 & 54.64 & 38.87 & 6.18 & 21.76 & 32.13 \\
            \midrule
            \textbf{Mistral-7B} & 59.50 & 52.69 & 50.12 & 39.96 & 19.64 & 40.76 & 27.91 & 42.53 & 40.77 & 16.82 & 24.87 & 28.67 & 33.77 & 29.11 & 69.92 & 54.42 & 85.65 & 58.59 & 31.33 & 68.71 & 42.73 \\
            \textbf{Mistral-7B (IT)} & 69.87 & 41.12 & 62.24 & 45.94 & 18.77 & 56.67 & 42.11 & 59.65 & 46.70 & 44.55 & 23.59 & 28.40 & 35.59 & 28.45 & 68.35 & 56.97 & 89.01 & 65.37 & 29.81 & 67.77 & 47.64 \\
            \midrule
            \textbf{Mixtral-8x7B} & 66.62 & 69.32 & 77.13 & 36.11 & 31.89 & 43.33 & 34.39 & 48.65 & 39.90 & 17.56 & 25.47 & 30.35 & 36.13 & 37.40 & 79.17 & 63.21 & 91.27 & 68.35 & 40.00 & 77.09 & 50.44 \\
            \textbf{Mixtral-8x7B (IT)} & 81.17 & 73.47 & 62.52 & 42.31 & 28.53 & 57.04 & 49.82 & 64.84 & 46.70 & 46.58 & 25.31 & 28.38 & 35.69 & 35.54 & 77.55 & 58.55 & 91.09 & 68.84 & 37.06 & 76.18 & 53.10 \\
            \midrule
            \textbf{Pythia-160M} & 51.59 & 45.57 & 51.35 & 72.44 & 0.02 & 52.51 & 24.72 & 46.48 & 26.99 & 8.69 & 3.29 & 2.72 & 6.06 & 0.11 & 0.28 & 50.32 & 50.16 & 37.96 & 0.11 & 0.28 & 20.00 \\
            \textbf{Pythia-410M} & 51.72 & 48.36 & 48.85 & 64.10 & 1.61 & 28.89 & 22.77 & 41.03 & 31.20 & 16.45 & 15.52 & 14.88 & 9.11 & 1.97 & 9.11 & 69.94 & 50.12 & 39.49 & 1.97 & 9.11 & 25.46 \\
            \textbf{Pythia-1B} & 45.27 & 51.59 & 49.82 & 54.38 & 2.47 & 27.66 & 23.99 & 40.48 & 32.63 & 17.38 & 18.95 & 15.58 & 11.42 & 2.77 & 14.07 & 72.05 & 50.34 & 39.22 & 2.77 & 14.07 & 26.50 \\
            \textbf{Pythia-1.4B} & 46.86 & 48.50 & 52.36 & 56.62 & 4.68 & 25.95 & 22.89 & 39.10 & 34.45 & 16.27 & 19.42 & 16.53 & 13.18 & 5.18 & 18.70 & 71.74 & 51.98 & 39.22 & 6.40 & 20.33 & 29.96 \\
            \textbf{Pythia-2.8B} & 45.00 & 42.44 & 55.38 & 50.96 & 6.00 & 23.62 & 21.05 & 36.27 & 34.74 & 16.45 & 20.99 & 19.39 & 20.05 & 7.95 & 27.41 & 70.95 & 51.98 & 44.43 & 9.06 & 30.11 & 30.86 \\
            \textbf{Pythia-6.9B} & 47.58 & 48.24 & 44.98 & 47.23 & 5.57 & 25.70 & 21.66 & 36.32 & 37.83 & 20.01 & 24.13 & 22.15 & 24.76 & 8.70 & 34.43 & 70.39 & 51.88 & 42.91 & 11.41 & 35.97 & 30.52 \\
            \textbf{Pythia-12B} & 49.65 & 45.29 & 54.94 & 42.41 & 7.73 & 25.70 & 20.56 & 32.04 & 37.99 & 18.11 & 22.72 & 20.77 & 21.73 & 10.61 & 39.77 & 67.31 & 54.41 & 42.14 & 12.69 & 41.60 & 32.98 \\
            \bottomrule
        \end{tabular}
    }
    \caption{Complete results on individual tasks.}
    \label{tab:all}
\end{table*}

\begin{table*}[ht!]
    \centering
    \resizebox{0.9\linewidth}{!}{
    % \begin{tabular}{lrrrrrr}
    \begin{tabular}{lcccccc}
        \toprule
         & Hallu. Detection & Instr. Following & QA & Reading Comp. & Sum. & Fact-Checking \\
        \midrule
        \textbf{Falcon-7B} & 52.40 & 25.54 & 33.03 & 29.18 & 20.70 & 46.91 \\
        \textbf{Falcon-7B (IT)} & 49.50 & 38.16 & 28.61 & 25.54 & 19.70 & 39.20 \\
        \midrule
        \textbf{FalconMamba-7B} & 55.80 & 42.97 & 39.94 & 23.76 & 23.89 & 61.85 \\
        \textbf{FalconMamba-7B (IT)} & 56.53 & 41.78 & 39.51 & 22.91 & 23.80 & 60.05 \\
        \midrule
        \textbf{Gemma-2B} & 53.01 & 30.94 & 27.43 & 30.35 & 19.63 & 43.28 \\
        \textbf{Gemma-7B} & 50.45 & 31.25 & 38.20 & 39.96 & 25.93 & 54.14 \\
        \textbf{Gemma-2B (IT)} & 49.74 & 40.73 & 23.30 & 27.52 & 23.67 & 38.06 \\
        \textbf{Gemma-7B (IT)} & 61.16 & 44.09 & 31.28 & 25.00 & 20.40 & 55.74 \\
        \textbf{Gemma1.1-2B (IT)} & 50.61 & 41.27 & 25.43 & 29.38 & 21.00 & 34.65 \\
        \textbf{Gemma1.1-7B (IT)} & 55.27 & 46.50 & 36.66 & 26.63 & 22.32 & 58.96 \\
        \midrule
        \textbf{Gemma2-2B} & 58.33 & 26.73 & 29.59 & 32.95 & 16.04 & 39.57 \\
        \textbf{Gemma2-9B} & 65.64 & 23.76 & 39.36 & 39.74 & 21.38 & 62.06 \\
        \textbf{Gemma2-27B} & 62.03 & 26.62 & 47.19 & 43.04 & 28.92 & 68.28 \\
        \textbf{Gemma2-2B (IT)} & 58.97 & 37.63 & 34.42 & 27.69 & 15.57 & 58.53 \\
        \textbf{Gemma2-9B (IT)} & 71.95 & 53.83 & 42.49 & 39.04 & 18.03 & 68.53 \\
        \textbf{Gemma2-27B (IT)} & 74.46 & 60.20 & 50.05 & 48.06 & 28.18 & 69.28 \\
        \midrule
        \textbf{RecurrentGemma-2B} & 53.68 & 34.43 & 27.36 & 31.66 & 19.41 & 42.97 \\
        \textbf{RecurrentGemma-9B} & 54.95 & 30.67 & 36.79 & 35.57 & 21.99 & 51.25 \\
        \textbf{RecurrentGemma-2B (IT)} & 55.21 & 36.55 & 25.51 & 28.89 & 21.27 & 46.60 \\
        \textbf{RecurrentGemma-9B (IT)} & 53.28 & 43.63 & 38.82 & 28.91 & 23.83 & 64.61 \\
        \midrule
        \textbf{LLaMA2-7B} & 53.63 & 28.94 & 37.64 & 27.58 & 25.19 & 51.38 \\
        \textbf{LLaMA2-13B} & 67.00 & 25.57 & 39.65 & 31.17 & 26.33 & 62.35 \\
        \textbf{LLaMA2-70B} & 62.34 & 30.24 & 47.85 & 39.48 & 28.00 & 66.63 \\
        \textbf{LLaMA2-7B (IT)} & 57.93 & 37.93 & 35.75 & 31.32 & 25.88 & 57.26 \\
        \textbf{LLaMA2-13B (IT)} & 64.92 & 37.29 & 39.00 & 29.53 & 28.10 & 60.71 \\
        \textbf{LLaMA2-70B (IT)} & 73.12 & 47.35 & 45.32 & 31.27 & 27.45 & 65.18 \\
        \midrule
        \textbf{LLaMA3-8B} & 60.76 & 22.06 & 41.55 & 33.52 & 26.62 & 60.84 \\
        \textbf{LLaMA3-70B} & 71.78 & 21.59 & 48.07 & 46.38 & 28.91 & 69.57 \\
        \textbf{LLaMA3-8B (IT)} & 70.88 & 48.82 & 42.85 & 37.14 & 25.02 & 65.66 \\
        \textbf{LLaMA3-70B (IT)} & 80.47 & 47.19 & 51.50 & 38.65 & 27.23 & 70.56 \\
        \midrule
        \textbf{Jamba} & 57.66 & 43.36 & 39.50 & 33.00 & 23.72 & 59.88 \\
        \textbf{Jamba (IT)} & 54.76 & 37.60 & 38.52 & 27.12 & 8.24 & 57.28 \\
        \midrule
        \textbf{Finch-1.6B} & 53.09 & 31.10 & 22.69 & 16.60 & 11.77 & 39.47 \\
        \textbf{Finch-3B} & 53.29 & 27.80 & 25.96 & 17.89 & 0.03 & 39.26 \\
        \textbf{Finch-7B} & 55.04 & 25.55 & 34.58 & 21.05 & 5.42 & 45.26 \\
        \textbf{Finch-14B} & 60.65 & 26.64 & 37.54 & 22.40 & 1.89 & 51.98 \\
        \midrule
        \textbf{Mamba-130M} & 55.24 & 44.29 & 13.31 & 12.87 & 12.84 & 33.84 \\
        \textbf{Mamba-370M} & 52.41 & 39.30 & 15.25 & 12.97 & 15.04 & 38.55 \\
        \textbf{Mamba-790M} & 50.81 & 35.35 & 18.14 & 18.75 & 16.96 & 38.73 \\
        \textbf{Mamba-1.4B} & 51.88 & 33.11 & 19.98 & 20.44 & 17.11 & 39.08 \\
        \textbf{Mamba-2.8B} & 53.10 & 33.74 & 23.91 & 20.80 & 20.00 & 42.99 \\
        \midrule
        \textbf{Mamba2-130M} & 54.02 & 46.13 & 14.05 & 12.73 & 13.88 & 39.17 \\
        \textbf{Mamba2-370B} & 53.41 & 38.58 & 16.19 & 19.26 & 17.05 & 39.08 \\
        \textbf{Mamba2-780B} & 52.97 & 36.58 & 17.42 & 21.44 & 19.15 & 38.35 \\
        \textbf{Mamba2-1.3B} & 53.77 & 34.63 & 19.45 & 22.20 & 17.98 & 41.26 \\
        \textbf{Mamba2-2.7B} & 53.62 & 31.40 & 22.82 & 27.90 & 20.66 & 40.35 \\
        \textbf{Mamba2Attn-2.7B} & 53.60 & 29.57 & 19.64 & 36.83 & 21.72 & 43.37 \\
        \midrule
        \textbf{Mistral-7B} & 60.48 & 28.39 & 41.24 & 32.03 & 26.77 & 58.59 \\
        \textbf{Mistral-7B (IT)} & 63.84 & 45.24 & 46.45 & 33.95 & 26.00 & 65.37 \\
        \midrule
        \textbf{Mixtral-8X7B} & 73.51 & 26.84 & 48.99 & 40.66 & 27.91 & 68.35 \\
        \textbf{Mixtral-8X7B (IT)} & 73.36 & 44.44 & 53.32 & 36.76 & 26.84 & 68.84 \\
        \midrule
        \textbf{Pythia-160M} & 49.80 & 40.56 & 10.42 & 8.14 & 3.00 & 37.96 \\
        \textbf{Pythia-410M} & 53.18 & 40.28 & 11.72 & 20.58 & 15.20 & 39.49 \\
        \textbf{Pythia-1B} & 53.81 & 35.88 & 13.78 & 27.11 & 17.26 & 39.22 \\
        \textbf{Pythia-1.4B} & 54.29 & 36.45 & 17.90 & 22.16 & 17.98 & 39.22 \\
        \textbf{Pythia-2.8B} & 53.15 & 33.71 & 20.18 & 22.89 & 20.19 & 44.43 \\
        \textbf{Pythia-6.9B} & 52.62 & 33.62 & 22.47 & 24.86 & 23.14 & 42.91 \\
        \textbf{Pythia-12B} & 54.32 & 30.26 & 23.84 & 28.05 & 21.75 & 42.14 \\
        \bottomrule
        \end{tabular}
    }
    \caption{Results separated by task categories.}
    \label{tab:categories}
\end{table*}

\begin{table*}[ht!]
    \centering
    \resizebox{0.9\linewidth}{!}{
    \begin{tabular}{lcccccc}
    \toprule
     & Hallu. Detection & Instr. Following & QA & Reading Comp. & Sum. & Fact-Checking \\
    \midrule
    \textbf{Falcon-7B} & 49.53 & 29.11 & 43.50 & 26.18 & 20.70 & 46.91 \\
    \textbf{Falcon-7B (IT)} & 45.22 & 44.21 & 29.13 & 21.13 & 19.71 & 39.20 \\
    \midrule
    \textbf{FalconMamba-7B} & 52.15 & 47.19 & 42.70 & 19.98 & 23.90 & 61.85 \\
    \textbf{FalconMamba-7B (IT)} & 52.20 & 45.63 & 41.77 & 19.44 & 23.80 & 60.05 \\
    \midrule
    \textbf{Gemma-2B} & 50.86 & 36.43 & 34.54 & 28.08 & 20.64 & 43.28 \\
    \textbf{Gemma-2B (IT)} & 49.29 & 43.60 & 19.19 & 23.57 & 21.71 & 38.06 \\
    \textbf{Gemma-7B} & 45.01 & 32.90 & 44.58 & 34.76 & 24.93 & 54.14 \\
    \textbf{Gemma-7B (IT)} & 60.66 & 47.33 & 29.69 & 23.04 & 20.41 & 55.74 \\
    \textbf{Gemma1.1-2B (IT)} & 50.52 & 44.41 & 21.27 & 26.53 & 20.02 & 34.65 \\
    \textbf{Gemma1.1-7B (IT)} & 51.32 & 47.12 & 37.28 & 26.31 & 22.32 & 58.96 \\
    \midrule
    \textbf{Gemma2-2B} & 55.10 & 31.01 & 37.82 & 32.13 & 16.04 & 39.57 \\
    \textbf{Gemma2-2B (IT)} & 54.83 & 40.62 & 35.33 & 28.15 & 15.57 & 58.53 \\
    \textbf{Gemma2-9B} & 62.73 & 26.61 & 49.30 & 38.02 & 21.35 & 62.06 \\
    \textbf{Gemma2-9B (IT)} & 71.21 & 53.89 & 47.89 & 37.78 & 18.02 & 68.53 \\
    \textbf{Gemma2-27B} & 58.23 & 29.05 & 58.76 & 40.52 & 28.90 & 68.28 \\
    \textbf{Gemma2-27B (IT)} & 74.87 & 57.08 & 57.62 & 44.46 & 28.17 & 69.28 \\
    \midrule
    \textbf{RecurrentGemma-2B} & 49.05 & 38.86 & 34.90 & 27.59 & 18.42 & 42.97 \\
    \textbf{RecurrentGemma-2B (IT)} & 51.02 & 40.49 & 25.93 & 27.11 & 21.28 & 46.60 \\
    \textbf{RecurrentGemma-9B} & 51.49 & 34.80 & 46.83 & 34.27 & 22.99 & 51.25 \\
    \textbf{RecurrentGemma-9B (IT)} & 50.33 & 47.66 & 45.24 & 29.00 & 24.83 & 64.61 \\
    \midrule
    \textbf{LLaMA2-7B} & 51.62 & 31.69 & 47.96 & 25.65 & 25.18 & 51.38 \\
    \textbf{LLaMA2-7B (IT)} & 55.95 & 41.30 & 40.15 & 29.34 & 25.89 & 57.26 \\
    \textbf{LLaMA2-13B} & 66.34 & 28.98 & 51.85 & 31.01 & 27.32 & 62.35 \\
    \textbf{LLaMA2-13B (IT)} & 62.10 & 37.64 & 46.84 & 27.44 & 27.10 & 60.71 \\
    \textbf{LLaMA2-70B} & 60.75 & 33.58 & 61.27 & 36.44 & 27.98 & 66.63 \\
    \textbf{LLaMA2-70B (IT)} & 74.25 & 49.63 & 52.59 & 30.43 & 27.43 & 65.18 \\
    \midrule
    \textbf{LLaMA3-8B} & 57.88 & 25.19 & 51.81 & 32.13 & 26.60 & 60.84 \\
    \textbf{LLaMA3-8B (IT)} & 71.37 & 51.59 & 49.53 & 35.39 & 25.02 & 65.66 \\
    \textbf{LLaMA3-70B} & 70.61 & 24.64 & 61.02 & 45.04 & 28.88 & 69.57 \\
    \textbf{LLaMA3-70B (IT)} & 81.70 & 45.77 & 59.18 & 37.54 & 27.21 & 70.56 \\
    \midrule
    \textbf{Jamba} & 53.44 & 50.30 & 43.45 & 29.15 & 23.72 & 59.88 \\
    \textbf{Jamba (IT)} & 50.28 & 47.46 & 38.92 & 25.42 & 8.23 & 57.28 \\
    \midrule
    \textbf{Finch-1.6B} & 51.43 & 35.07 & 24.94 & 10.35 & 11.73 & 39.47 \\
    \textbf{Finch-3B} & 51.38 & 30.74 & 30.36 & 14.33 & 0.03 & 39.26 \\
    \textbf{Finch-7B} & 53.22 & 27.29 & 41.06 & 20.56 & 5.39 & 45.26 \\
    \textbf{Finch-14B} & 58.69 & 28.23 & 43.98 & 23.51 & 1.88 & 51.98 \\
    \midrule
    \textbf{Mamba-130M} & 52.91 & 51.59 & 4.08 & 5.25 & 12.84 & 33.84 \\
    \textbf{Mamba-370M} & 49.43 & 44.82 & 9.81 & 6.45 & 15.04 & 38.55 \\
    \textbf{Mamba-790M} & 47.69 & 39.27 & 15.91 & 11.06 & 16.96 & 38.73 \\
    \textbf{Mamba-1.4B} & 48.63 & 36.97 & 20.59 & 12.91 & 17.12 & 39.08 \\
    \textbf{Mamba-2.8B} & 51.38 & 37.37 & 27.53 & 15.26 & 20.00 & 42.99 \\
    \midrule
    \textbf{Mamba2-130M} & 50.59 & 51.86 & 4.21 & 4.28 & 12.87 & 39.17 \\
    \textbf{Mamba2-370B} & 50.90 & 42.72 & 9.26 & 12.18 & 17.05 & 39.08 \\
    \textbf{Mamba2-780B} & 49.55 & 39.81 & 14.22 & 14.34 & 18.16 & 38.35 \\
    \textbf{Mamba2-1.3B} & 51.44 & 38.46 & 19.64 & 16.15 & 17.99 & 41.26 \\
    \textbf{Mamba2-2.7B} & 50.96 & 33.72 & 26.81 & 21.93 & 20.67 & 40.35 \\
    \textbf{Mamba2Attn-2.7B} & 49.56 & 32.84 & 24.12 & 30.08 & 21.72 & 43.37 \\
    \midrule
    \textbf{Mistral-7B} & 58.98 & 31.48 & 51.43 & 30.90 & 26.75 & 58.59 \\
    \textbf{Mistral-7B (IT)} & 62.47 & 45.43 & 50.98 & 31.94 & 25.98 & 65.37 \\
    \midrule
    \textbf{Mixtral-8X7B} & 73.44 & 29.32 & 60.87 & 38.99 & 27.89 & 68.35 \\
    \textbf{Mixtral-8X7B (IT)} & 74.02 & 43.87 & 59.64 & 34.24 & 26.83 & 68.84 \\
    \midrule
    \textbf{Pythia-160M} & 49.68 & 49.09 & 1.16 & 5.03 & 3.01 & 37.96 \\
    \textbf{Pythia-410M} & 51.34 & 46.65 & 7.76 & 16.13 & 15.20 & 39.49 \\
    \textbf{Pythia-1B} & 51.18 & 40.83 & 11.60 & 22.24 & 17.28 & 39.22 \\
    \textbf{Pythia-1.4B} & 51.67 & 41.84 & 14.72 & 15.96 & 17.99 & 39.22 \\
    \textbf{Pythia-2.8B} & 50.36 & 38.32 & 20.82 & 19.26 & 20.20 & 44.43 \\
    \textbf{Pythia-6.9B} & 47.25 & 34.26 & 24.80 & 19.11 & 20.14 & 42.91 \\
    \textbf{Pythia-12B} & 52.20 & 33.51 & 28.72 & 23.42 & 21.75 & 42.14 \\
    \bottomrule
    \end{tabular}
    }
    \caption{Results separated by task categories, where scores are weighted by the size of the tasks.}
    \label{tab:categories-weighted}
\end{table*}

\begin{table*}[ht!]
    \centering
    \resizebox{\linewidth}{!}{
    % \begin{tabular}{lrrrrrr}
    \begin{tabular}{lcccccccc}
        \toprule
        \textbf{Model Name} & \textbf{Hallu. Detection} & \textbf{Instr. Following} & \textbf{Closed-Book QA} & \textbf{Reading Comp.} & \textbf{Sum.} & \textbf{Fact-Checking} & \textsc{Faithfulness} & \textsc{Factuality} \\
        \midrule
        \multicolumn{9}{c}{\textit{Attention-Only Models}} \\
        \midrule
        \textbf{Gemma-2B} & 53.81 \da{\tt -5.67} & 31.94 \ua{\tt\pz 6.79} & 27.43 \da{\tt -4.13} & 31.35 \da{\tt\pz -5.83} & 20.63 \ua{\tt\pz 1.04} & 43.28  \da{\tt -5.22} & 34.32 \ua{\tt\pz 0.83} & 39.38 \da{\tt -11.20}\\
        \textbf{Gemma-7B} & 49.65 \ua{\tt 11.51} & 30.25 \ua{\tt 13.84} & 38.20 \da{\tt -6.92} & 38.96 \da{\tt -13.96} & 24.93 \da{\tt -4.53} & 54.14  \ua{\tt\pz 1.60} & 33.06 \ua{\tt\pz 4.92} & 50.44 \da{\tt\pz -8.42}\\
        \midrule
        \textbf{Gemma2-2B} & 58.33 \ua{\tt\pz 0.64} & 26.73 \ua{\tt 10.90} & 29.59 \ua{\tt\pz 4.83} & 32.95 \da{\tt -5.26} & 16.04 \da{\tt -0.47} & 39.57  \ua{\tt 18.96} & 33.97 \ua{\tt\pz 2.01} & 40.54 \ua{\tt\pz 6.17}\\
        \textbf{Gemma2-9B} & 65.64 \ua{\tt\pz 6.31} & 23.76 \ua{\tt 30.07} & 39.36 \ua{\tt\pz 3.13} & 39.74 \da{\tt -0.70} & 21.38 \da{\tt -3.35} & 62.06  \ua{\tt\pz 6.47} & 41.41 \ua{\tt\pz 3.93} & 56.42 \ua{\tt\pz 1.52}\\
        \textbf{Gemma2-27B} & 62.03 \ua{\tt 12.43} & 26.62 \ua{\tt 33.58} & 47.19 \ua{\tt\pz 2.86} & 43.04 \ua{\tt\pz 5.02} & 28.92 \da{\tt -0.74} & 68.28  \ua{\tt\pz 1.00} & 41.82 \ua{\tt 10.00} & 64.25 \da{\tt -0.31}\\
        \midrule
        \textbf{LLaMA2-7B} & 53.63 \ua{\tt\pz 4.30} & 28.94 \ua{\tt\pz 8.99} & 37.64 \da{\tt -1.89} & 27.58 \ua{\tt\pz 3.74} & 25.19 \ua{\tt\pz 0.69} & 51.38  \ua{\tt\pz 5.88} & 35.47 \ua{\tt\pz 2.90} & 50.65 \da{\tt -1.91}\\
        \textbf{LLaMA2-13B} & 67.80 \da{\tt -3.68} & 26.57 \ua{\tt\pz 9.72} & 39.65 \da{\tt -0.65} & 32.17 \da{\tt -3.64} & 27.33 \da{\tt -0.23} & 62.35  \da{\tt -1.64} & 41.08 \ua{\tt\pz 3.11} & 57.51 \da{\tt -3.38}\\
        \textbf{LLaMA2-70B} & 62.34 \ua{\tt 10.78} & 30.24 \ua{\tt 17.11} & 47.85 \da{\tt -2.53} & 39.48 \da{\tt -8.21} & 28.00 \da{\tt -0.55} & 66.63  \da{\tt -1.45} & 42.47 \ua{\tt\pz 5.96} & 64.76 \da{\tt -5.64}\\
        \midrule
        \textbf{LLaMA3-8B} & 60.76 \ua{\tt 10.12} & 22.06 \ua{\tt 26.76} & 41.55 \ua{\tt\pz 1.30} & 33.52 \ua{\tt\pz 3.62} & 26.62 \da{\tt -1.60} & 60.84  \ua{\tt\pz 4.82} & 39.32 \ua{\tt\pz 7.82}	& 57.22 \ua{\tt\pz 0.49}\\
        \textbf{LLaMA3-70B} & 71.78 \ua{\tt\pz 8.69} & 21.59 \ua{\tt 25.60} & 48.07 \ua{\tt\pz 3.43} & 46.38 \da{\tt -7.73} & 28.91 \da{\tt -1.68} & 69.57  \ua{\tt\pz 0.99} & 49.24 \ua{\tt\pz 4.00} & 66.05 \da{\tt -0.72} \\
        \midrule
        \textbf{Mistral-7B} & 60.48 \ua{\tt\pz 3.36} & 28.39 \ua{\tt 16.85} & 41.24 \ua{\tt\pz 5.21} & 32.03 \ua{\tt\pz 1.92} & 26.77 \da{\tt -0.77} & 58.59  \ua{\tt\pz 6.78} & 39.98 \ua{\tt\pz 1.85}	& 56.04 \ua{\tt\pz 2.40}\\
        \midrule
        \textbf{Mixtral-8X7B} & 73.51 \da{\tt -0.15} & 26.84 \ua{\tt 17.60} & 48.99 \ua{\tt\pz 4.33} & 40.66 \da{\tt -3.90} & 27.91 \da{\tt -1.07} & 68.35  \ua{\tt\pz 0.49} & 48.49 \ua{\tt\pz 0.82} & 65.37 \da{\tt -0.54}\\
        \midrule
        \textbf{Falcon-7B} & 52.40 \da{\tt -2.90} & 25.54 \ua{\tt 12.62} & 33.03 \da{\tt -4.42} & 29.18 \da{\tt -3.64} & 20.70 \da{\tt -1.00} & 46.91  \da{\tt -7.71} & 30.37 \ua{\tt\pz 3.89}	& 45.60 \da{\tt -10.78}\\
        \midrule
        \multicolumn{9}{c}{\textit{Recurrent and Hybrid Models}} \\
        \midrule
        \textbf{RecurrentGemma-2B} & 52.88 \ua{\tt\pz 2.33} & 33.43 \ua{\tt\pz 3.12} & 27.36 \da{\tt -1.85} & 30.66 \da{\tt -1.77} & 18.41 \ua{\tt\pz 2.86} & 42.97  \ua{\tt\pz 3.63} & 34.07 \da{\tt -1.11} & 38.93 \da{\tt -2.45}\\
        \textbf{RecurrentGemma-9B} & 55.75 \da{\tt -1.67} & 31.67 \ua{\tt 12.96} & 36.79 \ua{\tt\pz 2.03} & 36.57 \da{\tt -6.66} & 22.99 \ua{\tt\pz 1.84} & 51.25  \ua{\tt 13.36} & 35.89 \da{\tt -1.29} & 50.63 \ua{\tt\pz 4.11} \\
        \midrule
        \textbf{Jamba} & 57.66 \da{\tt -2.90} & 43.36 \da{\tt -5.76} & 39.50 \da{\tt -0.98} & 33.00 \da{\tt -5.88} & 23.72 \da{\tt -15.48} & 59.88 \da{\tt -2.60} & 36.28 \da{\tt -7.38} & 51.78 \da{\tt -3.49}\\
        \midrule
        \textbf{FalconMamba-7B} & 55.80 \ua{\tt\pz 0.73} & 42.97 \da{\tt -1.19} & 39.94 \da{\tt -0.43} & 23.76 \da{\tt -0.85} & 23.89 \da{\tt -0.09} & 61.85 \da{\tt -1.80} & 33.31 \ua{\tt\pz 0.28} & 52.37 \da{\tt -1.11}\\
        \bottomrule
        \end{tabular}
    }
    \caption{Changes in performance from the use of instruction-tuning, weighted by the number of samples present in each task.}
    \label{tab:instruction_tuning-table-weighted}
\end{table*}

\begin{table}[ht!]
    \centering
    \resizebox{0.95\linewidth}{!}{
        % \begin{tabular}{lrr}
        \begin{tabular}{lcc}
            \toprule
             & \textbf{Faithfulness} & \textbf{Factuality} \\
            \midrule
            \textbf{Falcon-7B} & 34.91 & 36.92 \\
            \textbf{Falcon-7B (IT)} & 34.20 & 33.17 \\
            \midrule
            \textbf{FalconMamba-7B} & 35.92 & 47.02 \\
            \textbf{FalconMamba-7B (IT)} & 35.66 & 46.64 \\
            \midrule
            \textbf{Finch-1.6B} & 31.62 & 27.41 \\
            \textbf{Finch-3B} & 29.21 & 30.14 \\
            \textbf{Finch-7B} & 30.52 & 38.65 \\
            \textbf{Finch-14B} & 32.08 & 42.69 \\
            \midrule
            \textbf{Gemma-2B} & 35.85 & 32.56 \\
            \textbf{Gemma-7B} & 36.48 & 44.50 \\
            \textbf{Gemma-2B (IT)} & 36.77 & 28.00 \\
            \textbf{Gemma-7B (IT)} & 39.19 & 38.46 \\
            \textbf{Gemma1.1-2B (IT)} & 36.60 & 30.12 \\
            \textbf{Gemma1.1-7B (IT)} & 37.11 & 43.79 \\
            \midrule
            \textbf{Gemma2-2B} & 36.61 & 34.96 \\
            \textbf{Gemma2-9B} & 40.56 & 46.85 \\
            \textbf{Gemma2-27B} & 41.54 & 53.92 \\
            \textbf{Gemma2-2B (IT)} & 36.26 & 41.94 \\
            \textbf{Gemma2-9B (IT)} & 48.11 & 49.98 \\
            \textbf{Gemma2-27B (IT)} & 54.56 & 56.28 \\
            \midrule
            \textbf{RecurrentGemma-2B} & 36.55 & 31.46 \\
            \textbf{RecurrentGemma-2B (IT)} & 36.41 & 32.85 \\
            \textbf{RecurrentGemma-9B} & 36.61 & 42.96 \\
            \textbf{RecurrentGemma-9B (IT)} & 36.44 & 46.24 \\
            \midrule
            \textbf{Jamba} & 39.80 & 45.95 \\
            \textbf{Jamba (IT)} & 32.89 & 45.04 \\
            \midrule
            \textbf{LLaMA2-7B} & 35.14 & 42.51 \\
            \textbf{LLaMA2-13B} & 40.83 & 46.27 \\
            \textbf{LLaMA2-70B} & 41.74 & 53.69 \\
            \textbf{LLaMA2-7B (IT)} & 39.15 & 42.39 \\
            \textbf{LLaMA2-13B (IT)} & 41.71 & 45.78 \\
            \textbf{LLaMA2-70B (IT)} & 47.56 & 51.35 \\
            \midrule
            \textbf{LLaMA3-8B} & 37.60 & 48.14 \\
            \textbf{LLaMA3-70B} & 45.92 & 54.80 \\
            \textbf{LLaMA3-8B (IT)} & 47.69 & 49.74 \\
            \textbf{LLaMA3-70B (IT)} & 52.15 & 57.61 \\
            \midrule
            \textbf{Mamba-130M} & 34.38 & 19.13 \\
            \textbf{Mamba-370M} & 32.56 & 21.21 \\
            \textbf{Mamba-790M} & 32.95 & 23.63 \\
            \textbf{Mamba-1.4B} & 33.44 & 25.23 \\
            \textbf{Mamba-2.8B} & 34.47 & 29.05 \\
            \midrule
            \textbf{Mamba2-130M} & 34.42 & 20.13 \\
            \textbf{Mamba2-370B} & 34.93 & 22.04 \\
            \textbf{Mamba2-780B} & 35.40 & 22.90 \\
            \textbf{Mamba2-1.3B} & 35.31 & 24.91 \\
            \textbf{Mamba2-2.7B} & 36.39 & 27.86 \\
            \textbf{Mamba2Attn-2.7B} & 38.91 & 26.49 \\
            \midrule
            \textbf{Mistral-7B} & 38.47 & 47.42 \\
            \textbf{Mistral-7B (IT)} & 43.14 & 52.59 \\
            \midrule
            \textbf{Mixtral-8X7B} & 46.16 & 55.15 \\
            \textbf{Mixtral-8X7B (IT)} & 48.05 & 58.65 \\
            \midrule
            \textbf{Pythia-160M} & 25.13 & 15.23 \\
            \textbf{Pythia-410M} & 33.29 & 18.69 \\
            \textbf{Pythia-1B} & 35.43 & 18.21 \\
            \textbf{Pythia-1.4B} & 35.89 & 23.44 \\
            \textbf{Pythia-2.8B} & 35.48 & 25.79 \\
            \textbf{Pythia-6.9B} & 36.30 & 27.46 \\
            \textbf{Pythia-12B} & 36.85 & 28.73 \\
            \bottomrule
        \end{tabular}
    }
    \caption{Results categorized by faithfulness and factuality.}
    \label{tab:group}
\end{table}

\begin{table}[ht!]
    \centering
    \resizebox{0.95\linewidth}{!}{
        \begin{tabular}{lrr}
        \toprule
         & \textbf{Faithfulness} & \textbf{Factuality} \\
        \midrule
        \textbf{Falcon-7B} & 30.37 & 45.60 \\
        \textbf{Falcon-7B (IT)} & 34.26 & 34.82 \\
        \midrule
        \textbf{FalconMamba-7B} & 32.75 & 52.37 \\
        \textbf{FalconMamba-7B (IT)} & 33.03 & 51.26 \\
        \midrule
        \textbf{Finch-1.6B} & 29.36 & 31.87 \\
        \textbf{Finch-3B} & 26.54 & 35.01 \\
        \textbf{Finch-7B} & 29.70 & 43.97 \\
        \textbf{Finch-14B} & 31.41 & 48.68 \\
        \midrule
        \textbf{Gemma-2B} & 34.32 & 39.38 \\
        \textbf{Gemma-2B (IT)} & 35.15 & 28.18 \\
        \textbf{Gemma-7B} & 33.06 & 50.44 \\
        \textbf{Gemma-7B (IT)} & 37.98 & 42.02 \\
        \textbf{Gemma1.1-2B (IT)} & 34.55 & 28.59 \\
        \textbf{Gemma1.1-7B (IT)} & 33.85 & 48.00 \\
        \midrule
        \textbf{Gemma2-2B} & 33.97 & 40.54 \\
        \textbf{Gemma2-2B (IT)} & 35.98 & 46.71 \\
        \textbf{Gemma2-9B} & 41.41 & 56.42 \\
        \textbf{Gemma2-9B (IT)} & 45.34 & 57.94 \\
        \textbf{Gemma2-27B} & 41.82 & 64.25 \\
        \textbf{Gemma2-27B (IT)} & 51.82 & 63.94 \\
        \midrule
        \textbf{RecurrentGemma-2B} & 34.07 & 38.93 \\
        \textbf{RecurrentGemma-2B (IT)} & 32.96 & 36.48 \\
        \textbf{RecurrentGemma-9B} & 35.89 & 50.63 \\
        \textbf{RecurrentGemma-9B (IT)} & 34.60 & 54.74 \\
        \midrule
        \textbf{Jamba} & 36.28 & 51.78 \\
        \textbf{Jamba (IT)} & 28.90 & 48.29 \\
        \midrule
        \textbf{LLaMA2-7B} & 35.47 & 50.65 \\
        \textbf{LLaMA2-7B (IT)} & 38.37 & 48.74 \\
        \textbf{LLaMA2-13B} & 41.08 & 57.51 \\
        \textbf{LLaMA2-13B (IT)} & 44.19 & 54.13 \\
        \textbf{LLaMA2-70B} & 42.47 & 64.76 \\
        \textbf{LLaMA2-70B (IT)} & 48.43 & 59.12 \\
        \midrule
        \textbf{LLaMA3-8B} & 39.32 & 57.22 \\
        \textbf{LLaMA3-8B (IT)} & 47.14 & 57.71 \\
        \textbf{LLaMA3-70B} & 49.24 & 66.05 \\
        \textbf{LLaMA3-70B (IT)} & 53.24 & 65.33 \\
        \midrule
        \textbf{Mamba-130M} & 29.78 & 17.64 \\
        \textbf{Mamba-370M} & 28.72 & 22.70 \\
        \textbf{Mamba-790M} & 29.28 & 26.35 \\
        \textbf{Mamba-1.4B} & 30.14 & 29.23 \\
        \textbf{Mamba-2.8B} & 32.78 & 34.83 \\
        \midrule
        \textbf{Mamba2-130M} & 28.44 & 19.58 \\
        \textbf{Mamba2-370B} & 31.37 & 22.59 \\
        \textbf{Mamba2-780B} & 31.50 & 25.16 \\
        \textbf{Mamba2-1.3B} & 32.76 & 29.41 \\
        \textbf{Mamba2-2.7B} & 34.30 & 33.42 \\
        \textbf{Mamba2Attn-2.7B} & 34.45 & 33.51 \\
        \midrule
        \textbf{Mistral-7B} & 39.98 & 56.04 \\
        \textbf{Mistral-7B (IT)} & 41.83 & 58.44 \\
        \midrule
        \textbf{Mixtral-8X7B} & 48.49 & 65.37 \\
        \textbf{Mixtral-8X7B (IT)} & 49.31 & 64.83 \\
        \midrule
        \textbf{Pythia-160M} & 22.83 & 12.62 \\
        \textbf{Pythia-410M} & 30.48 & 18.94 \\
        \textbf{Pythia-1B} & 32.71 & 20.29 \\
        \textbf{Pythia-1.4B} & 32.92 & 25.81 \\
        \textbf{Pythia-2.8B} & 33.60 & 31.22 \\
        \textbf{Pythia-6.9B} & 31.84 & 32.97 \\
        \textbf{Pythia-12B} & 35.72 & 35.11 \\
        \bottomrule
        \end{tabular}
    }
    \caption{Results categorized by faithfulness and factuality, weighted by the number of examples per task.}
    \label{tab:group-weighted}
\end{table}

% \subsection{Instruction Tuning Changes}

\begin{figure*}[ht!]
    \centering
    \caption{Change in task performance from base model to instruction fine-tuned model, for all tasks.}
    \includegraphics[width=\linewidth]{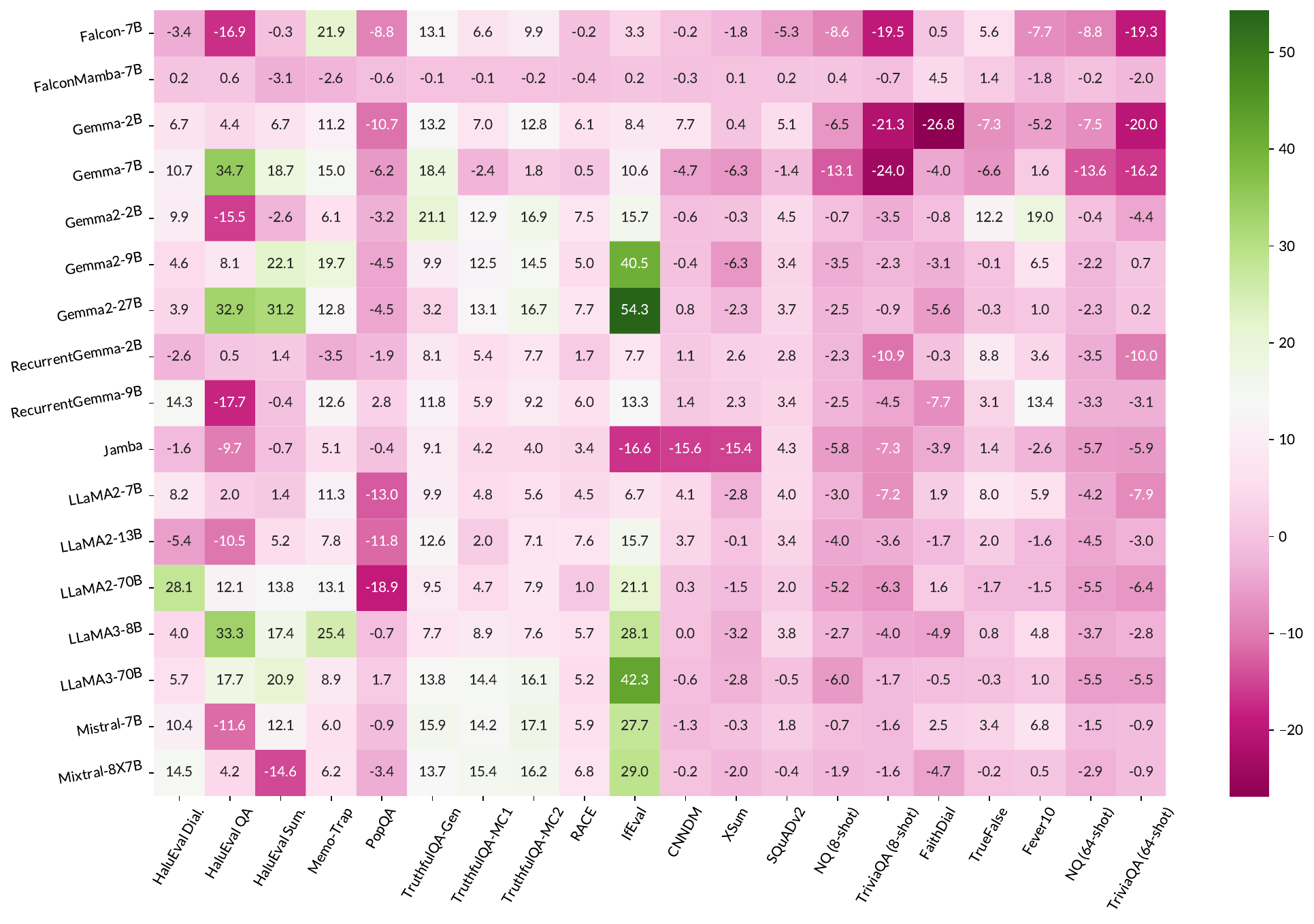}
    \label{fig:instruction_tuning_change-tasks-figure}
    \vspace{-0.5cm}
    % \caption{Change in overall task category performance from base to instruction fine-tuned model.}
    % \includegraphics[width=\linewidth]{figures/categories-changes.pdf}
    % \label{fig:instruction_tuning_change-categories-figure}
\end{figure*}

\begin{figure*}[ht!]
    \centering
    \caption{Change in task category performance from base model to instruction fine-tuned model.}
    \includegraphics[width=\linewidth]{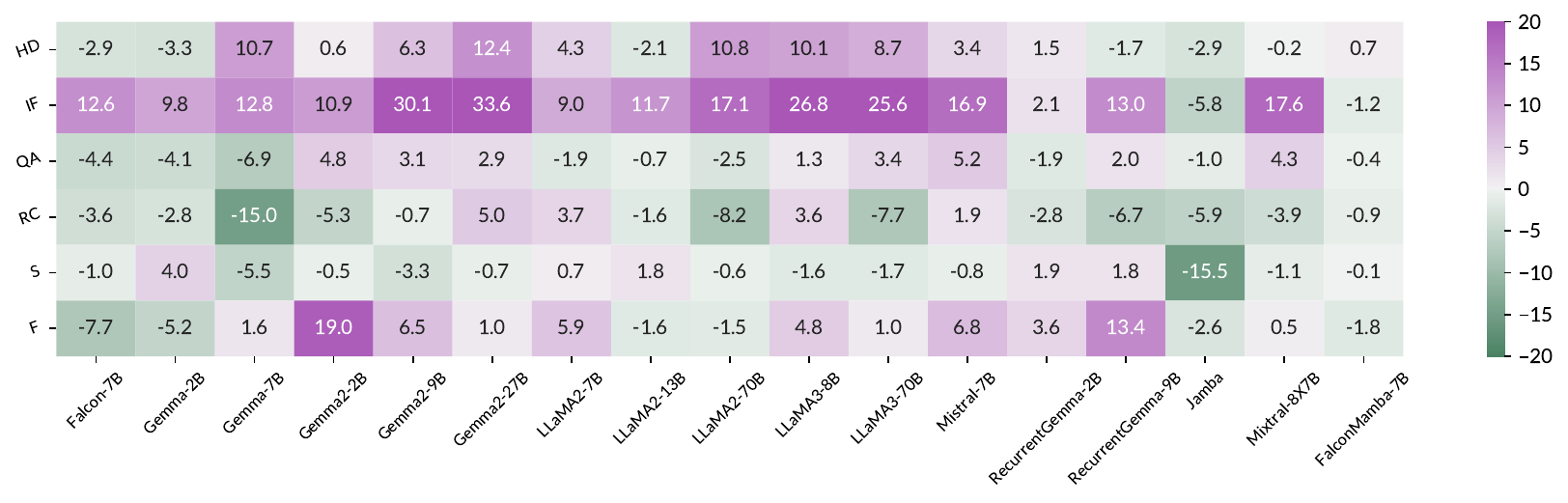}
    \label{fig:instruction_tuning_change-categoreis-figure}
    \vspace{-0.5cm}
    % \caption{Change in overall task category performance from base to instruction fine-tuned model.}
    % \includegraphics[width=\linewidth]{figures/categories-changes.pdf}
    % \label{fig:instruction_tuning_change-categories-figure}
\end{figure*}

\end{document}